\PassOptionsToPackage{table}{xcolor}
\documentclass[twocolumn,switch]{article}
\usepackage{preprint}
\usepackage[utf8]{inputenc}
\usepackage[T1]{fontenc}
\usepackage{amsmath,amsfonts}
\usepackage{algorithmic}
\usepackage{algorithm}
\usepackage{array}
\usepackage{url}
\usepackage{graphicx}
\usepackage{enumitem}
\usepackage[section]{placeins}
\usepackage{booktabs}
\usepackage{multirow}
\usepackage{tabularx}
\usepackage{amssymb}
\usepackage{makecell}
\usepackage{siunitx}
\usepackage{textgreek}
\usepackage{csquotes}
\usepackage{pifont}
\usepackage{bm}
\usepackage{microtype}
\usepackage{dblfloatfix}
\newcommand{\cmark}{\ding{51}}
\newcommand{\xmark}{\ding{55}}
\usepackage[labelformat=simple]{subfig}

\usepackage[hidelinks]{hyperref}
\usepackage[capitalize]{cleveref}
\usepackage[numbers,sort&compress]{natbib}

\definecolor{darkred}{rgb}{0.6, 0, 0}
\definecolor{darkgreen}{rgb}{0, 0.5, 0}

\newcommand{\colorcm}{{\color{darkgreen}\cmark}}
\newcommand{\colorxm}{{\color{darkred}\xmark}}

\crefname{figure}{Fig.}{Figs.}
\newcommand{\ignore}[1]{}
\title{PCFootprint: A Large-Scale Dataset and Benchmark for Vectorized Building Footprint Extraction from Aerial LiDAR Point Clouds}

\author{
Haoyuan Shen\textsuperscript{$\dagger$}\and
Kuihao Wang\textsuperscript{$\dagger$}\and
Ruisheng Wang\textsuperscript{$*$}\and
Yujun Liu\textsuperscript{$*$}
}

\date{
School of Architecture and Urban Planning, Shenzhen University, Shenzhen 518060, China
}

\begin{document}

\twocolumn[
  \begin{@twocolumnfalse}
\maketitle
\thispagestyle{empty}

\begin{abstract}
Building footprint extraction is a fundamental task in photogrammetry, remote sensing, and computer vision. Recent image-based methods have achieved remarkable progress in extracting vectorized footprints from high-resolution optical imagery. However, optical imagery inherently susceptible to occlusions, perspective distortions, and residual relief displacement, yielding incomplete or misaligned footprint extraction. Furthermore, the lack of explicit elevation information limits its direct applicability to Level of Detail building modeling. In this paper, we present PCFootprint, the first large-scale public dataset for footprint extraction from airborne laser scanning point clouds. PCFootprint comprises \num{33000} tiles derived from the Estonian Land and Spatial Development Board, covering diverse urban and rural landscapes. Each tile spans \qtyproduct{128 x 128}{\m} with systematically aligned vectorized footprints aligned to point clouds. The dataset includes a \num{3000} tiles cross-domain test set for evaluating generalization across geographic regions. We establish comprehensive benchmarks by evaluating mainstream methods. Experimental results reveal significant challenges including high intra-class variance, data imbalance, and noise across complex geospatial environments. We believe PCFootprint will advance future research in building modeling, urban scene understanding, and geospatial analysis. The PCFootprint dataset is publicly available at \url{https://huggingface.co/datasets/Haoyuan-Shen/PCFootprint}.
\end{abstract}

\vspace{0.35cm}

  \end{@twocolumnfalse}
]

\section{Introduction}
{\let\thefootnote\relax\footnotetext{\textsuperscript{$\dagger$}Equal contribution. \textsuperscript{$*$}Corresponding authors. \par {This work has been submitted to the IEEE for possible publication. Copyright may be transferred without notice, after which this version may no longer be accessible.}}}

Building footprint extraction serves as a core information creation and applied remote sensing process, standing at the intersection of photogrammetry, remote sensing, and computer vision to empower digital twin and smart city infrastructures~\cite{Zhu2019MAP-Net,li_hd-net_2024,Li2019Semantic,Bittner2018Building}.
Building footprints are typically represented in two formats: pixel-wise rasterized masks or vectorized polygons defined by ordered vertices. However, buildings exhibit diverse architectural styles, complex structures, and varied geometric configurations across different urban scenes. Therefore, automated and accurate extraction of building footprints remains a challenging and important research problem. 

Recent advances in image-based datasets and methods have achieved remarkable progress in vectorized footprint extraction~\cite{Wei2020Toward,wei_buildmapper_2023,wang_sampolybuild_2024,zhang2024p2pformer,yu2025p2pformerv2,jiao_roipoly_2025,adimoolam2025pix2poly,wang2025holitracer}. Large-scale datasets such as WHU Building~\cite{ji_WHU_2019}, Inria~\cite{maggiori_inria_2017}, and CrowdAI~\cite{mohanty2020crowdai} provide high-resolution optical imagery for extracting building geometries. These datasets have enabled methods to produce precise footprint boundaries under optimal imaging conditions. Despite these advances, image-based extraction faces inherent limitations that constrain practical applications. First, optical imagery is susceptible to spectral interference from shadows and occlusions from overhanging vegetation~\cite{Guo2022A,Yu2025Enhancing,Yang2025Advances}. These factors often obscure building edges and result in incomplete or fragmented extractions. Second, extracted footprints lack elevation information, which limits their direct use in Level of Detail 1 (LoD1) modeling~\cite{2024FusionHeightNet,Xu2024Building,Yu2024_3D}.

Airborne Laser Scanning (ALS) has emerged as a primary source for large-scale urban 3D data acquisition~\cite{Tomljenovic2015Building}. By providing point clouds with precise 3D coordinates, absolute scale, and penetration capability~\cite{Zang2024Compound,You2021Tree}, ALS enables reliable geometric representation of complex urban and rural environments and has been widely adopted for 3D perception, understanding, and reconstruction tasks~\cite{Domínguez2025Mapping,Li2020A}. However, a public benchmark dataset for extracting vectorized building footprints from ALS data is still lacking, which significantly hinders the development and systematic evaluation of point cloud-based footprint extraction methods. Existing large-scale 3D datasets~\cite{varney2020dales, han_whu-urban3d_2024}, primarily focus on point-wise semantic or instance labeling, and thus fail to provide structured, vectorized footprint representations required by Geographic Information Systems (GIS) applications, including spatial analysis and urban planning. Moreover, most existing footprint extraction studies in point cloud domain rely on private datasets, limiting reproducibility, fair performance comparison, and generalization across diverse urban scenes~\cite{awrangjeb_automatic_2014,gamal_semi-automatic_2023,nalini_automatic_2025,Kong2022Automatic,Kong2023PH-shape:,Nurunnabi2022ROBUST,Hrutka2022VOXEL-BASED}. Therefore, a standardized and large-scale ALS dataset dedicated to building footprint extraction is imperative. Such a benchmark would facilitate accurate boundary delineation in cluttered urban landscapes and occluded dense forested areas. As demonstrated in \cref{fig:task_define}, the inherent elevation information within the point clouds was exploited for the seamless LoD1 reconstruction, and further serve as reliable boundary constraints to improve LoD2 building modeling accuracy~\cite{huang2022city3d,Kong2025Large-Scale,Zang2024Compound,Yang2016Automated,Nurunnabi2022ROBUST}.  

In this paper, we present PCFootprint, the first large-scale public dataset for vectorized building footprint extraction from ALS point clouds. The data is derived from the Estonian Land and Spatial Development Board. It encompasses diverse urban and rural landscapes across multiple regions. The geographical distribution of these acquired point cloud tiles across Estonia is visualized in \cref{fig:pointcloud_tiles_sampling_geometry}. We initially selected \num{500} large point cloud tiles characterized by significant height variations and structural complexity. These tiles exhibit varying point densities across different geographic regions. The \num{500} tiles were subdivided into \num{30000} smaller point cloud tiles for processing. We provide an additional cross-domain generalization test set comprising \num{3000} point cloud tiles. Building instances fractured by tile boundaries were identified and removed from this test set. Ultimately, \num{33000} point cloud tiles were obtained for the complete dataset. Each tile covers a standard spatial extent of \qtyproduct{128 x 128}{\m}. We further establish a comprehensive benchmark for building footprint extraction from ALS point clouds and evaluate a wide range of mainstream methods. The results reveal substantial challenges arising from intra-class variation, data imbalance, and noise across complex geospatial environments. We believe PCFootprint will support future research in building modeling, 3D reconstruction, urban scene understanding, and geospatial analysis.

The primary contributions of this work are summarized as follows:

\begin{itemize}
    \item We present PCFootprint, the first large-scale dataset dedicated to the vectorized extraction of building footprints from ALS point clouds, which facilitates the automated city-scale LoD1 building reconstruction.
    \item The dataset features unprecedented scale with nationwide coverage across Estonia, comprising \num{33000} tiles with systematically aligned vectorized annotations for \num{227264} building instances to ensure national representativeness.
    \item We establish a standardized benchmarking platform featuring robust intra-domain evaluation and cross-domain generalization protocols, providing a verified baseline through the assessment of mainstream algorithms.
    \item Our open-source release of the PCFootprint benchmark bridges a critical gap in high-quality public resources, fostering reproducible research and community-wide advancement within the geospatial domain.
\end{itemize}

\section{Related Works}

\subsection{Building Footprint Extraction Datasets}
Building footprint extraction benchmarks are categorized by input modality and structural annotation format. Early research framed building footprint extraction as a pixel-wise semantic segmentation task. This paradigm is exemplified by datasets such as  Massachusetts~\cite{MnihThesis}, ISPRS~\cite{isprs_semantic_labeling}, and Inria~\cite{maggiori_inria_2017}, which represent ground truth in the form of rasterized binary masks. However, these rasterized representations frequently suffer from aliasing artifacts and irregular boundaries~\cite{wei_buildmapper_2023}, hindering their seamless integration with GIS workflows. Transforming these raster masks into regularized polygons necessitates computationally intensive post-processing.
To address these geometric limitations, a subsequent generation of datasets has pivoted toward native vectorized polygon annotations. This transition enables the development of end-to-end building vectorization and contour-based extraction methods that directly regress building corner vertices. For instance, the SpaceNet challenge series~\cite{Etten2018SpaceNetAR} released a corpus of high-resolution satellite imagery featuring over \num{685000} annotated polygonal building footprints. Similarly, the  WHU Building~\cite{ji_WHU_2019} and the WHU-Mix (Vector)~\cite{wei_buildmapper_2023} established robust benchmarks for footprint extraction by providing high-fidelity vector labels for large-scale aerial and satellite imagery. Despite their structural advantages, these image-centric frameworks remain fundamentally susceptible to spectral interference, relief displacement, and physical obstructions such as overhanging vegetation. Such optical artifacts frequently lead to fragmented extractions and inaccurate boundary delineations, particularly in cluttered urban environments~\cite{vostikolaei_multimodal_2024}.

Geometric fidelity and occlusion handling represent the inherent advantages of 3D point clouds over 2D imagery. Optical sensors often suffer from vegetation obstructions and shadow artifacts that blur semantic boundaries~\cite{Guo2022A,Dong2023A}. In contrast, 3D LiDAR systems feature multi-return and side-view acquisition. These capabilities enables the direct capture of underlying structures by effectively penetrating upper-level canopies~\cite{Raj2020A}. Moreover, unlike standard orthophotos, which are processed derivatives susceptible to residual relief displacement and perspective distortions, 3D sensing preserves absolute spatial coordinates. Although modern deep learning attempts to generate True Digital Orthophoto Maps (TDOMs) without explicit Digital Surface Model (DSM) priors~\cite{chen2025orthonerf,wang2025high_quality,wang2024tortho_gaussian}, these methods remain reliant on implicit radiometric consistency to rectify displacements. Consequently, direct 3D sensing remains indispensable for achieving the structural integrity required in GIS and digital twin frameworks.

Despite their high quality, prominent 3D urban datasets like DALES~\cite{varney2020dales}, WHU-Urban3D~\cite{han_whu-urban3d_2024}, and Toronto-3D~\cite{tan2020toronto} are predominantly optimized for point-wise semantic labeling or instance segmentation. While these benchmarks enable the classification of discrete points into categories such as \enquote{building} or \enquote{vegetation,} they lack the structured 2D polygon vectors necessary for direct geometric modeling. Consequently, research focusing on vectorized footprint extraction from LiDAR point clouds often resorts to localized, private, or small-scale collections such as the Aitkenvale or Hervey Bay collections~\cite{awrangjeb_automatic_2014}, which offer limited building instances. Other datasets rely on manually refined national inventories or datasets restricted to specific regions like Fredericton~\cite{vostikolaei_multimodal_2024}, Depok~\cite{gamal_semi-automatic_2023}, or Hyderabad~\cite{nalini_automatic_2025}. The lack of a standardized, large-scale public benchmark for building footprint extraction from ALS persists as a critical bottleneck, hindering both the objective assessment of geometric fidelity and the development of domain-invariant algorithms for complex 3D environments.
\begin{figure*}[t]
    \centering
    \subfloat[Nationwide raw building density of Estonia\label{fig:estonia_building_density}]{
        \includegraphics[width=0.48\textwidth, trim=15mm 5mm 15mm 15mm]{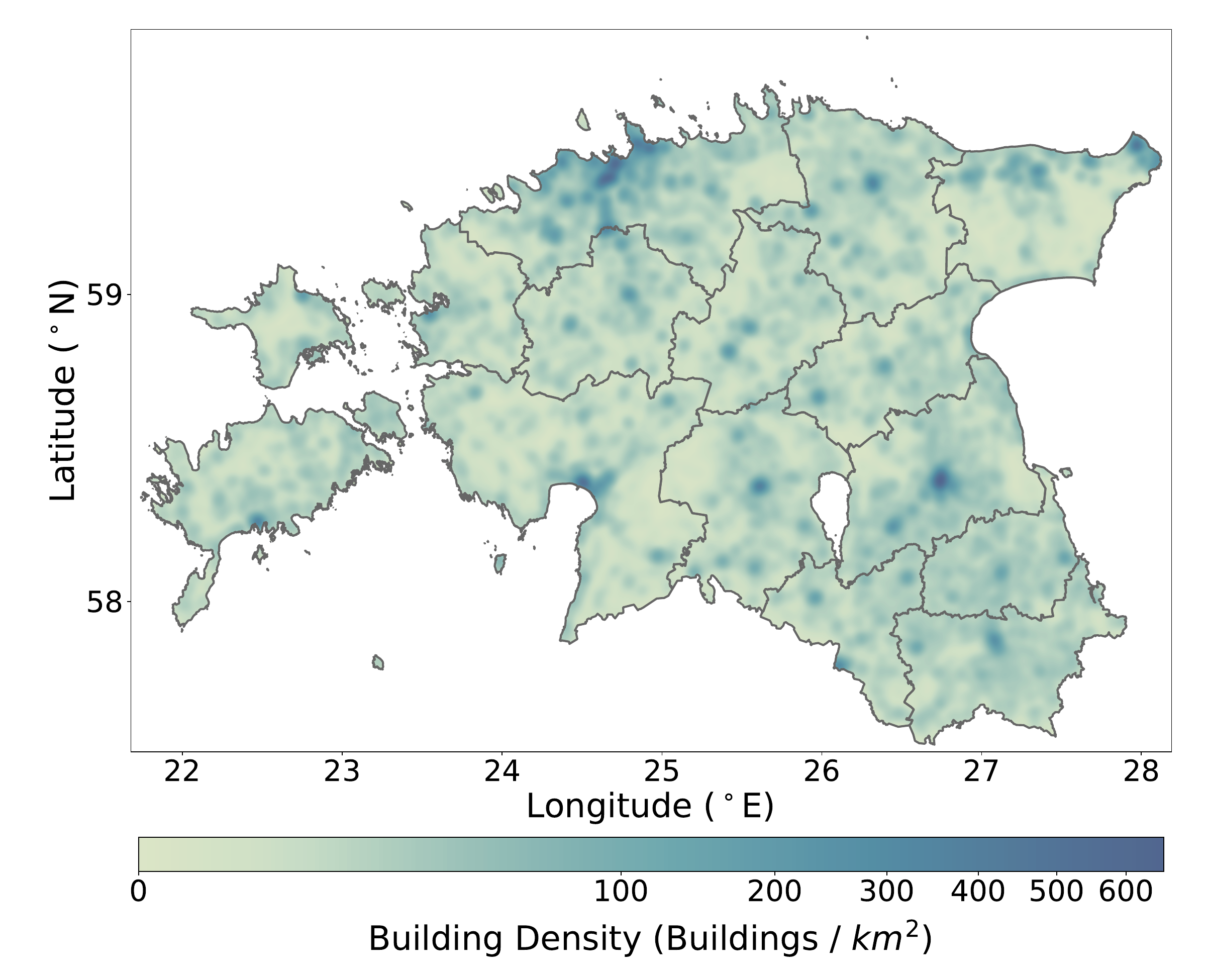}
    }
    \hfill
    \subfloat[Spatial distribution of sampled PCFootprint tiles\label{fig:pointcloud_tiles_sampling_geometry}]{
        \includegraphics[width=0.48\textwidth, trim=15mm 5mm 15mm 15mm]{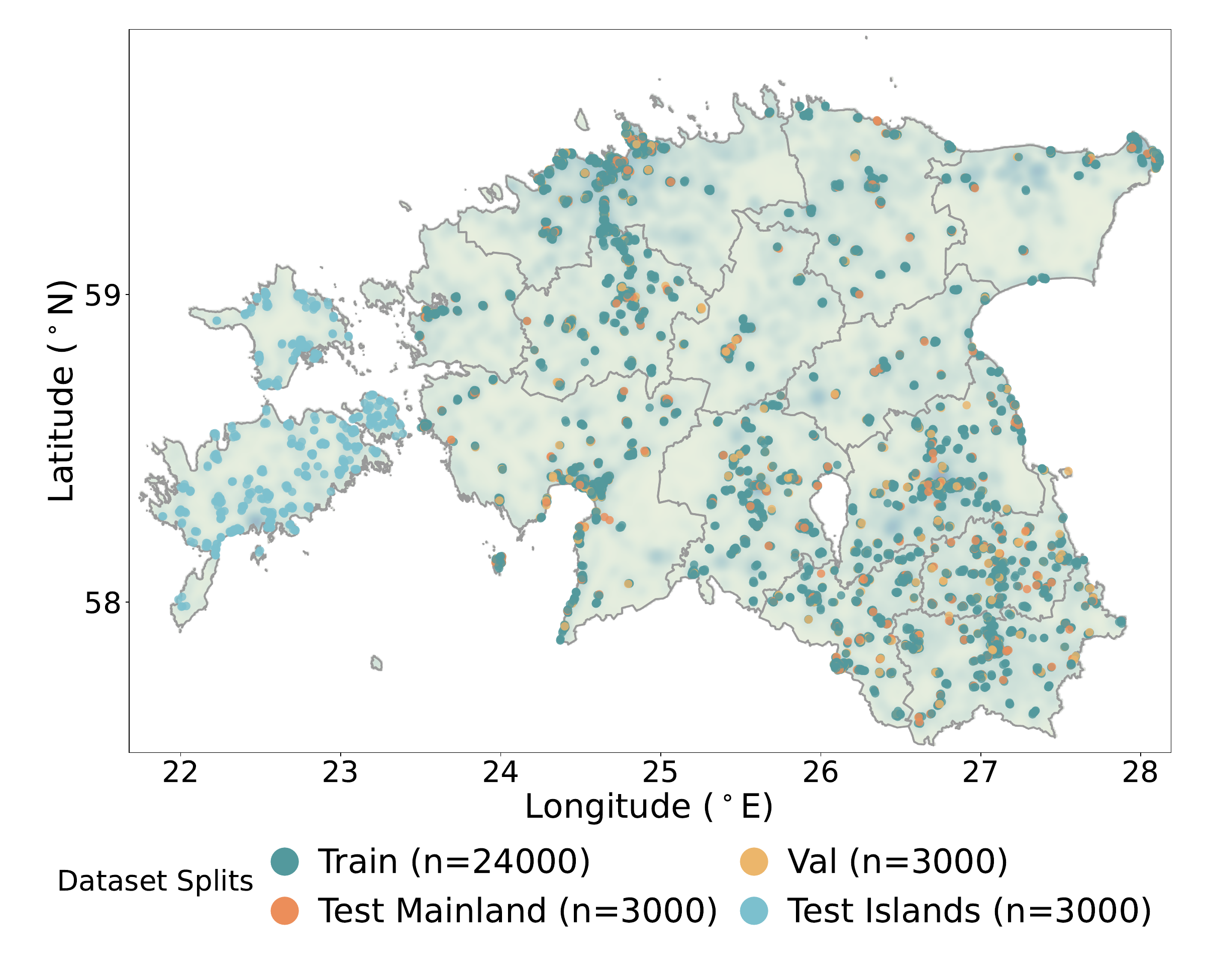}
    }
    \\
    \subfloat[Cumulative building instance statistics by counties\label{fig:building_counts_geometry}]{
        \includegraphics[width=0.48\textwidth, trim=15mm 5mm 15mm 15mm]{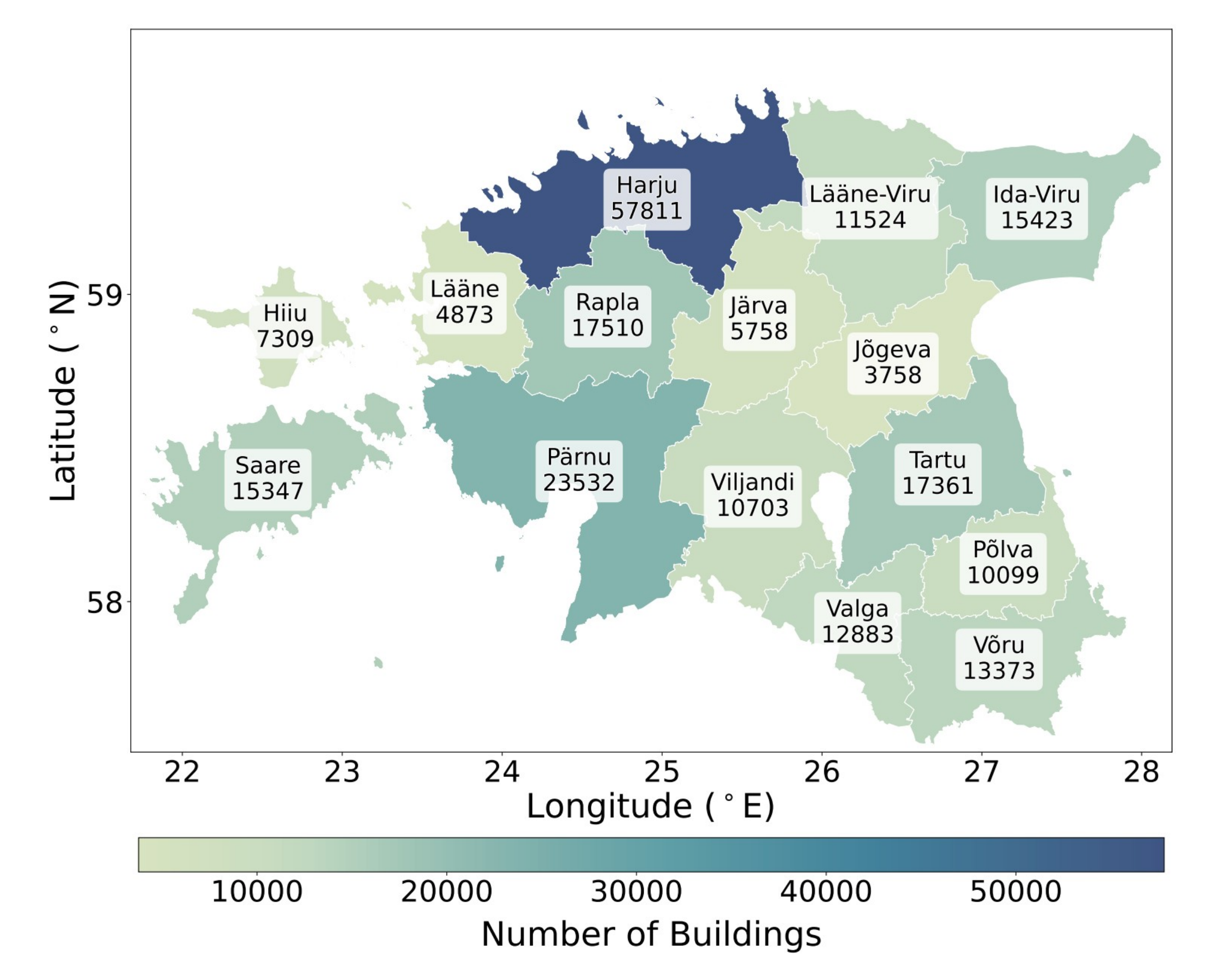}
    }
    \hfill
    \subfloat[Illustrative visualization of structured data representations\label{fig:task_define}]{
        \includegraphics[width=0.48\textwidth, trim=13.5mm 3.5mm 13.5mm 18mm]{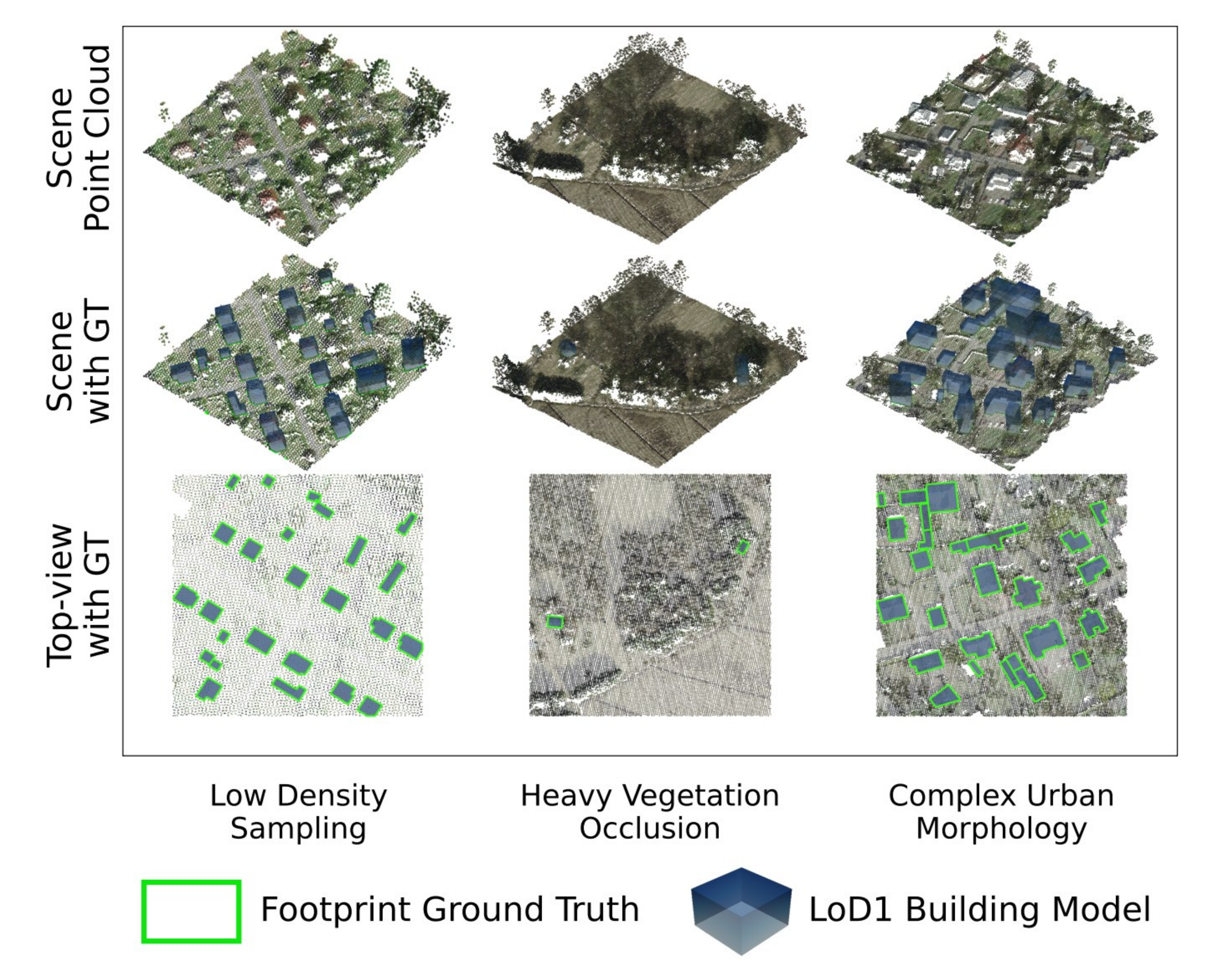}
    }
    
    \caption{Geographic sampling strategy and illustrative visualization of the PCFootprint dataset. (a) Nationwide building density heatmap of Estonia, serving as the geographic prior for data acquisition. (b) Spatial distribution of sampled point cloud tiles, which is strategically designed to be highly correlated with the actual building density shown in (a). (c) Cumulative building instances aggregated by count, demonstrating the dataset's massive scale, and validating that our sampling strategy accurately reflects real-world geographic patterns across Estonia. (d) Illustrative visualization of structured data representations, establishing the strong spatial correspondence between raw ALS point clouds (top), geographically aligned 3D LoD1 models (middle), and structured 2D vectorized footprints (bottom).}
    \label{fig:dataset_overview}
\end{figure*}
\subsection{Building Footprint Extraction Methodologies}
Current methods for building footprint extraction, largely driven by the availability of optical benchmarks, typically follow two primary paradigms: pixel-wise segmentation and polygonal regression. Pixel-wise segmentation approaches leverage the strong generalizability of Vision Foundation Models (VFMs) like SAM~\cite{kirillov2023sam}. Within this paradigm, MSA-SAM~\cite{chen2025msa-sam} addresses object scale variations, while SegEarth-OV3~\cite{li2025segearthov3} introduces open-vocabulary capabilities. Other advancements have integrated dynamic dictionary learning~\cite{zou2025dynamic}, style mixing for domain-invariant feature learning~\cite{luo_diverse_2023}, diffusion-based generative modeling~\cite{han2025cityinsight}, and frequency-guided structural encoding~\cite{Yuan2025FDENet} to refine extraction quality. However, the inherently discrete nature of these pixel-wise masks often yields irregular boundaries and lacks the sharp geometric topology required for high-precision mapping. To mitigate raster artifacts such as staircase effect, polygonal regression methods bypass intermediate rasterization by directly predicting vertex sequences or geometric primitives. Architectures such as P2PFormer~\cite{zhang2024p2pformer}, Pix2Poly~\cite{adimoolam2025pix2poly}, and HoliTracer~\cite{wang2025holitracer} leverage global Transformer-based sequence prediction or holistic tracing to capture topological relationships. In contrast, methods driven by target detection like PolyR-CNN~\cite{jiao2024polyrcnn} and RoIPoly~\cite{jiao_roipoly_2025} focus on efficient local vertex regression within Regions of Interest (RoIs), while other works achieve GIS-compatible outputs through contour evolution~\cite{wei_buildmapper_2023} or bottom-up line-primitive reconstruction~\cite{wei_lines_2024}.
Furthermore, SAM's segmentation power is integrated into polygonal regression paradigm via SAMPolyBuild~\cite{wang_sampolybuild_2024}. Nevertheless, these purely 2D-driven approaches lack intrinsic 3D geometric awareness, making them susceptible to spatial misalignments caused by varying terrain elevations. To bridge the gap between 2D imagery and 3D space, multi-modal frameworks such as MFNet~\cite{ma2025_mfnet} incorporate DSM elevation data into unified fine-tuning architectures to compensate for optical deficiencies. Despite these sophisticated advancements, image-based workflows remain fundamentally constrained by perspective distortions and the geometric information loss incurred during the projection process.

To overcome these limitations, methodologies operating directly within the 3D domain hold immense promise for ensuring absolute geometric fidelity and resolving complex occlusions. By exploiting raw spatial coordinates, 3D-centric approaches can theoretically achieve superior robustness against the shadows and perspective distortions that typically compromise 2D orthophotos. Research in this field, however, is currently dominated by traditional geometric heuristics such as LasBuildSeg~\cite{erdem2023_lasbuildseg}. Alongside other classical algorithms~\cite{rottmann2022automatic,awrangjeb_automatic_2014,nalini_automatic_2025,Nurunnabi2022ROBUST,Kong2023PH-shape:}, these methods rely on plane fitting and morphological operations that frequently necessitate manual parameter tuning. Recent deep learning attempts have addressed these issues by introducing terrain-aware self supervised learning to learn structural features from LiDAR data~\cite{vats_terrain-informed_2024}. While this approach enhances structural representation learning, it primarily generates pixel-level segmentation masks that lack direct vectorized structure and require extensive post-processing. Similarly, GAN-based frameworks have been proposed for parameter free footprint extraction from gridded images to eliminate manual tuning~\cite{Kong2022Automatic}. However, these methods still rely on intermediate raster representations rather than achieving end-to-end 3D vectorization. Such direct extraction of vectorized building footprints from point clouds is primarily stifled by a lack of large scale benchmarks with high-quality annotations. To bridge this gap, this paper introduces PCFootprint, the first extensive dataset for building footprint extraction from ALS designed to support the automated generation of structured, high-fidelity building footprints directly from 3D space.

\section{The PCFootprint Dataset}
PCFootprint introduces the first extensive, open-source benchmark specifically engineered for the vectorized extraction of building footprints from ALS point clouds. Sourced from the Estonian Land and Spatial Development Board, the dataset spans a vast geographic extent of \qty{540.67}{\km\squared}, comprising \num{33000} standardized tiles and \num{227264} systematically aligned building instances across diverse urban and rural landscapes. The detailed specifications of the dataset are summarized in \cref{tab:tab_dataset_profile}.
\begin{table}[htbp]
\centering
\small
\caption{Detailed specifications of the proposed PCFootprint dataset.}
\label{tab:tab_dataset_profile}
\renewcommand{\arraystretch}{1.3} 
\begin{tabularx}{0.8\columnwidth}{>{\raggedright\arraybackslash\bfseries}X >{\centering\arraybackslash}X} 
\toprule
Property & Details \\ 
\midrule
Subsets & Mainland, Islands \\
Data Type & ALS Point Cloud \\
Annotation Format & Vectorized Polygon \\
Accessibility & Open-source \\
Point Spacing [\unit{\m}] & \numrange{0.22}{1.09} \\
Tile Size & \qtyproduct{128 x 128}{\m} \\
Total Tiles & \num{33000} \\
Building Instances & \num{227264} \\
Coverage [\unit{\km\squared}] & \num{540.67} \\
\bottomrule
\end{tabularx}
\end{table}

\subsection{Dataset Construction}
The sampling strategy of PCFootprint is meticulously curated to reflect the geospatial diversity of Estonian urban and rural landscapes, fundamentally driven by the actual geospatial distribution of buildings to ensure national representativeness. We initially acquired \num{500} original tiles on the Estonia mainland for standard benchmarking workflows, each covering a \qtyproduct{1 x 1}{\km} area. To align with prevailing footprint extraction algorithms and optimize computational efficiency, each large tile was subdivided into \num{64} smaller tiles with a \qtyproduct{128 x 128}{\m} ground size, yielding a total of \num{30000} valid mainland samples. Following an identical processing pipeline, an additional \num{3000} tiles were acquired from the major islands of Saaremaa and Hiiumaa to support cross-domain evaluation. As demonstrated by the correlation between the nationwide building density heatmap (\cref{fig:estonia_building_density}) and the spatial allocation of tiles (\cref{fig:pointcloud_tiles_sampling_geometry}), the distribution of these \num{33000} tiles directly aligns with building concentration. This ensures that the dataset captures a representative cross-section topography and a diverse range of architectural styles. This statistical alignment is further supported by a closed validation of the cumulative instance counts in \cref{fig:building_counts_geometry}, confirming that the dataset reflects the real-world density patterns of Estonian counties.

To illustrate the diverse data characteristics captured through this geographically informed sampling, \cref{fig:task_define} presents an illustrative visualization of representative scene alongside their high fidelity structural annotations. These examples showcase a broad spectrum of data diversity within PCFootprint, ranging from fluctuating point densities and heavy vegetation occlusion to complex scene compositions with highly intricate building morphologies. By incorporating such geospatial heterogeneity, PCFootprint facilitates the rigorous evaluation of model robustness against varying land use patterns and enables researchers to better address the inherent challenges of cross-regional generalization. To ensure compatibility with mainstream deep learning architectures, all annotations adhere to the Microsoft Common Objects in Context (MS-COCO)~\cite{lin2014microsoft_coco} format. Each footprint is represented as a vectorized polygon defined by an ordered sequence of corner vertices, $V={\{v_1,~v_2,~\cdots,~v_n\}}$, where each vertex maps to a precise spatial coordinate within the ALS point cloud. As shown in \cref{fig:task_define}, this establishes a rigorous structural correspondence between the raw 3D input and the final structured 2D vectorized footprints.

Beyond horizontal geometry, the dataset incorporates a density-connectivity climbing procedure to calculate accurate building heights. By analyzing the vertical point density distribution, this approach effectively mitigates height distortions induced by sensor noise and vegetation occlusion. These calculated elevations subsequently serve as a rigorous criterion for data refinement. Specifically, instances with implausible height attributes were systematically removed, as such discrepancies typically stem from temporal inconsistencies between the LiDAR acquisition and the official footprint updates. Furthermore, to maintain structural integrity, we eliminated building instances intersected by tile boundaries. In such cases, both the vector annotations and their associated point clusters were discarded to preserve the quality of the remaining samples.

\subsection{Dataset Statistics}
This subsection provides a detailed quantitative characterization of the PCFootprint dataset, emphasizing the structural, geometric, and geospatial diversity that underpins the benchmark. To demonstrate the representativeness of the sampled data, statistics are organized across \num{15} administrative regions, capturing a broad spectrum of urban and rural profiles.

\begin{figure}[htbp] 
    \centering
    \includegraphics[width=\columnwidth]{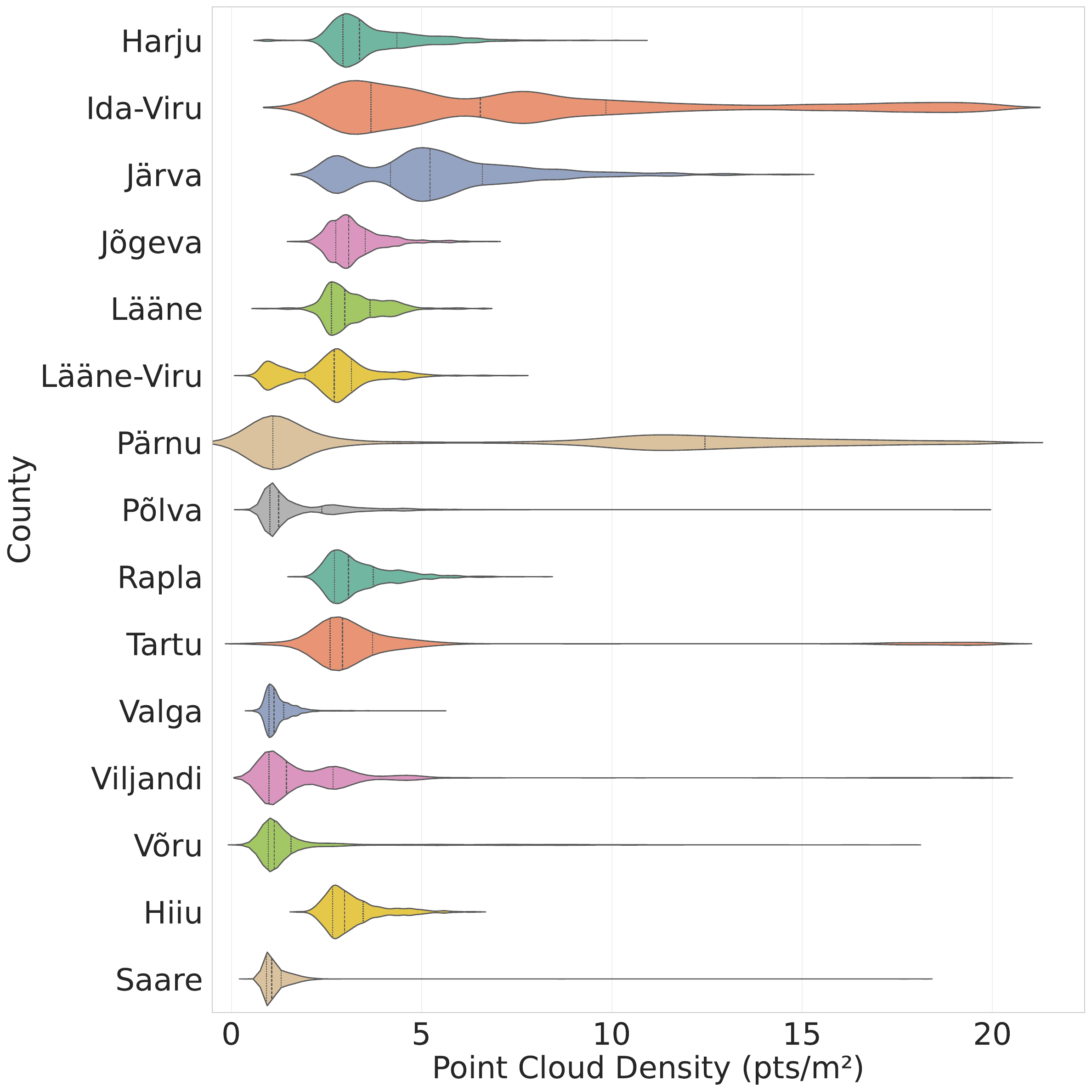} 
    \caption{Comparison of point cloud densities ($\mathrm{pts/m^2}$) across different counties in Estonia. Each violin plot shows the distribution of point cloud density, revealing the spread and concentration of points over building areas within each region. The multi-modal nature of these distributions indicates significant variations in LiDAR scanning patterns.}
    \label{fig:pointcloud_density_violin_p95}
\end{figure}

\textbf{Point Cloud Density Variations.} A critical challenge in extracting building footprints from ALS lies in the substantial heterogeneity of point cloud density and quality. As shown in the density distribution analysis in \cref{fig:pointcloud_density_violin_p95}, the point density within PCFootprint ranges from \qty{0.9}{pts\per\square\metre} to \qty{20.9}{pts\per\square\metre}. This variance is primarily attributed to different flight altitudes and sensor configurations used across various geographic regions. High-density areas provide rich geometric details for boundary localization, whereas low-density regions present substantial challenges for the detection of small structures and the preservation of topological integrity. By covering such a wide density range, this benchmark enables rigorous evaluation of algorithm robustness against non-uniform data quality.
\begin{figure}[htbp]
    \centering
    \includegraphics[width=\columnwidth]{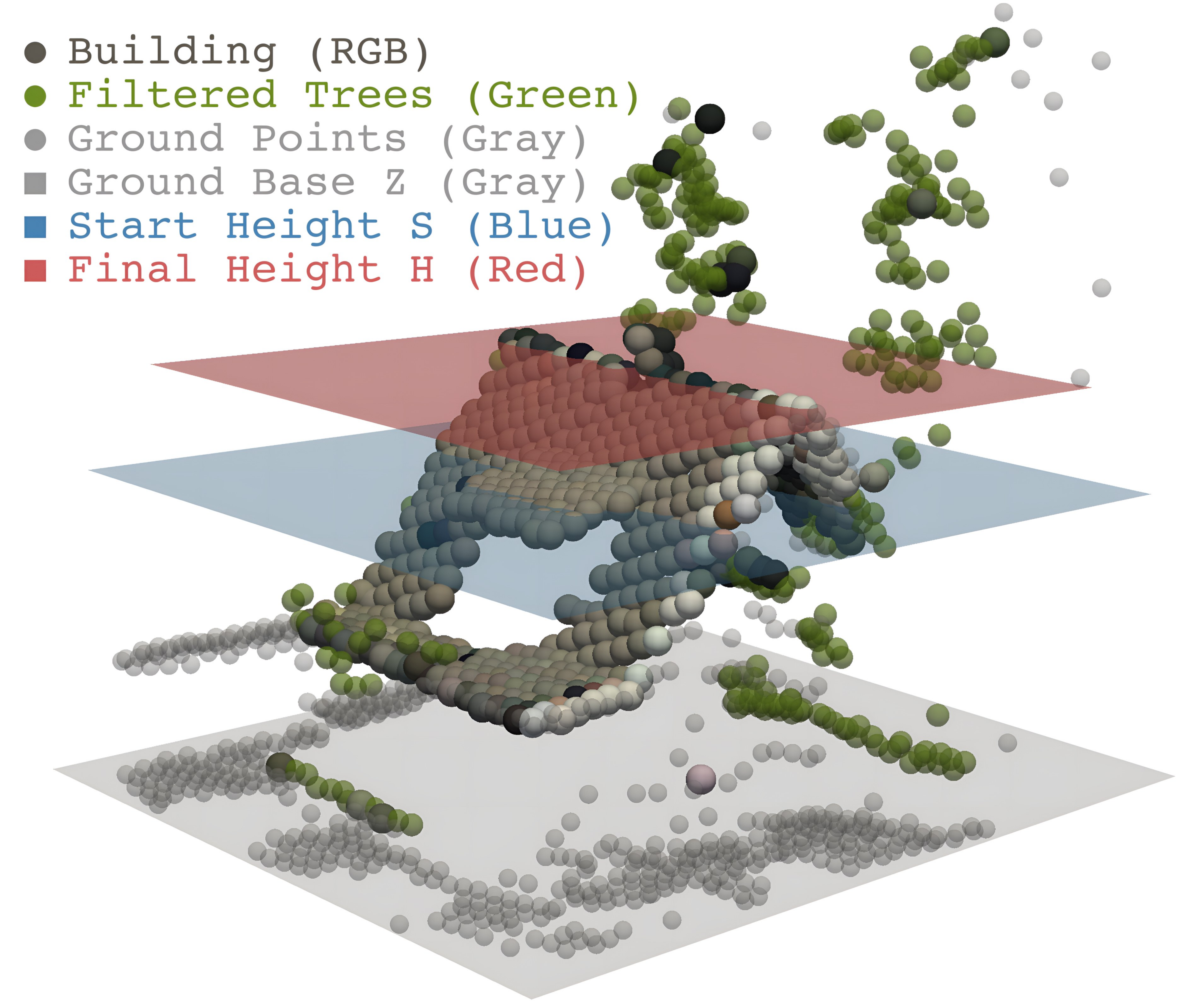}
    \caption{Conceptual workflow of the building height estimation. Following the suppression of vegetation interference (green points), the algorithm establishes a robust ground reference ($Z$) from filtered terrain returns (grey points). The primary building structure is then anchored at $S$ (blue plane) via hierarchical peak detection, with the final height ($H$, red plane) subsequently determined through density connectivity climbing.}
    \label{fig:height_calc_demo}
\end{figure}

\begin{figure}[htbp]
    \centering
    \includegraphics[width=\columnwidth]{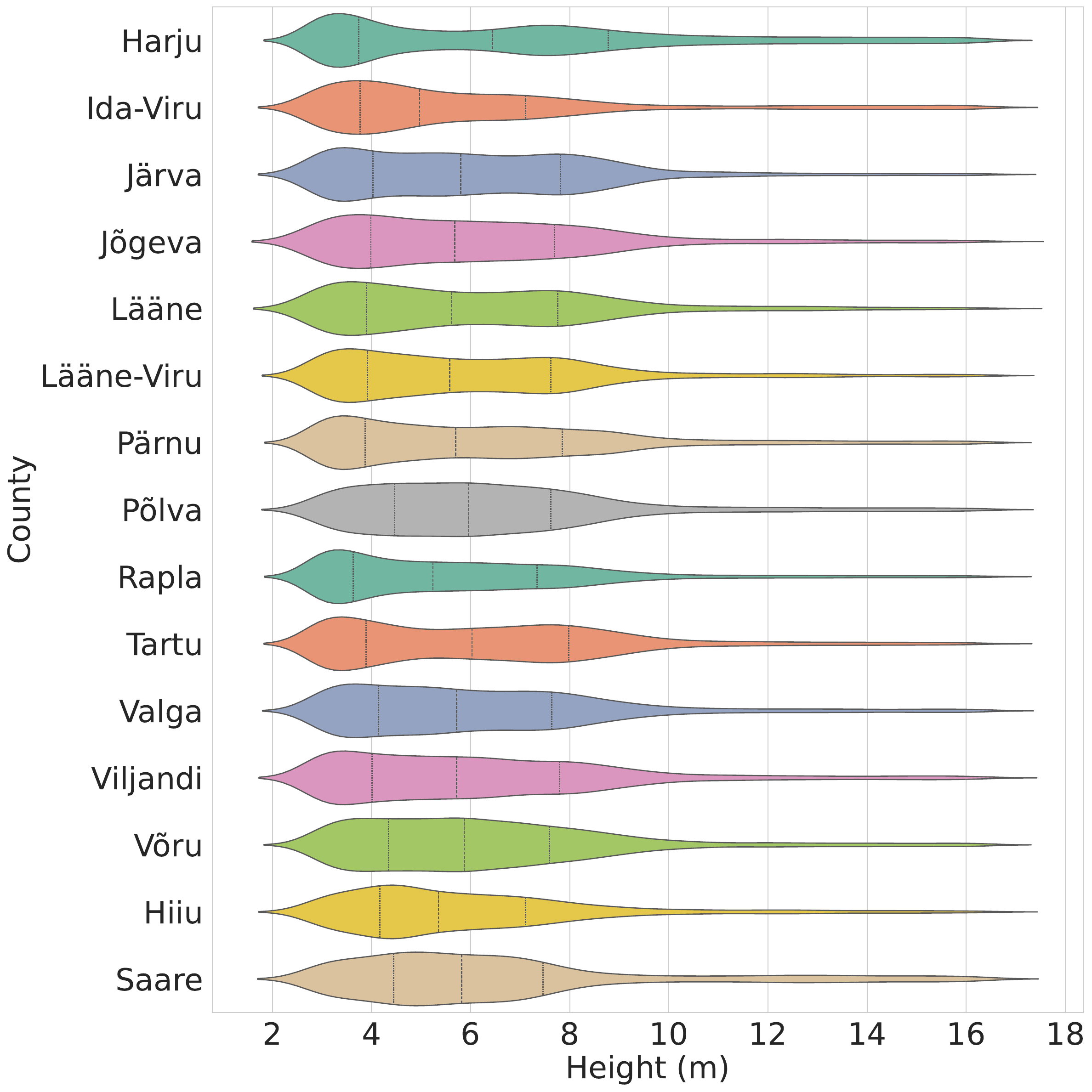}
    \caption{Statistical distribution of building heights across Estonian counties. The violin plots characterize the predominant building scales by focusing on the height distribution up to the $95^\text{th}$ percentile ($P_{95}$). Although the maximum height among all building instances reaches \qty{118.82}{\m}, the visualization is optimized to exclude these extreme outliers.}
    \label{fig:building_height_violin_p95}
\end{figure}

\textbf{Multi-scale Characteristics. } 
The structural complexity of PCFootprint is characterized by both vertical elevation and horizontal physical area. Regarding vertical dimensions, we eschew the naive use of maximum elevation within a footprint boundary to estimate building heights, which is often biased by sensor noise and vegetation interference. As demonstrated in \cref{fig:height_calc_demo}, we first filter vegetation interference (green points) using laser echo characteristics. To establish a robust reference plane $(Z)$, we then expand the horizontal search radius to incorporate surrounding ground returns. Subsequently, building heights are estimated via a density-connectivity climbing procedure. This approach identifies the primary starting height $(S)$ by detecting local point density peaks within a sliding vertical window. By exploiting the vertical density continuity of the building structure, the procedure iteratively climbs to update the elevation, terminating at a significant density discontinuity to determine the final building height $(H)$.
The vertical profiles in \cref{fig:building_height_violin_p95} faithfully reflect the geographic diversity of Estonia. Harju County, the urbanized core, exhibits a prominent concentration of tall buildings, evidenced by its elongated upper tail. Conversely, multi-peak distributions in Järva, Lääne, and Tartu signify a structural mix of low-rise residential and large industrial complexes. Notably, the striking vertical consistency between the islands (Saare, Hiiu) and mainland counties (e.g., Põlva) demonstrates the feasibility for cross-domain generalization. Such inherent height heterogeneity ensures that models learn a broad spectrum of vertical features, essential for high-fidelity 3D extraction across varied landscapes.
\begin{figure}[htbp]
    \centering
    \includegraphics[width=\columnwidth]{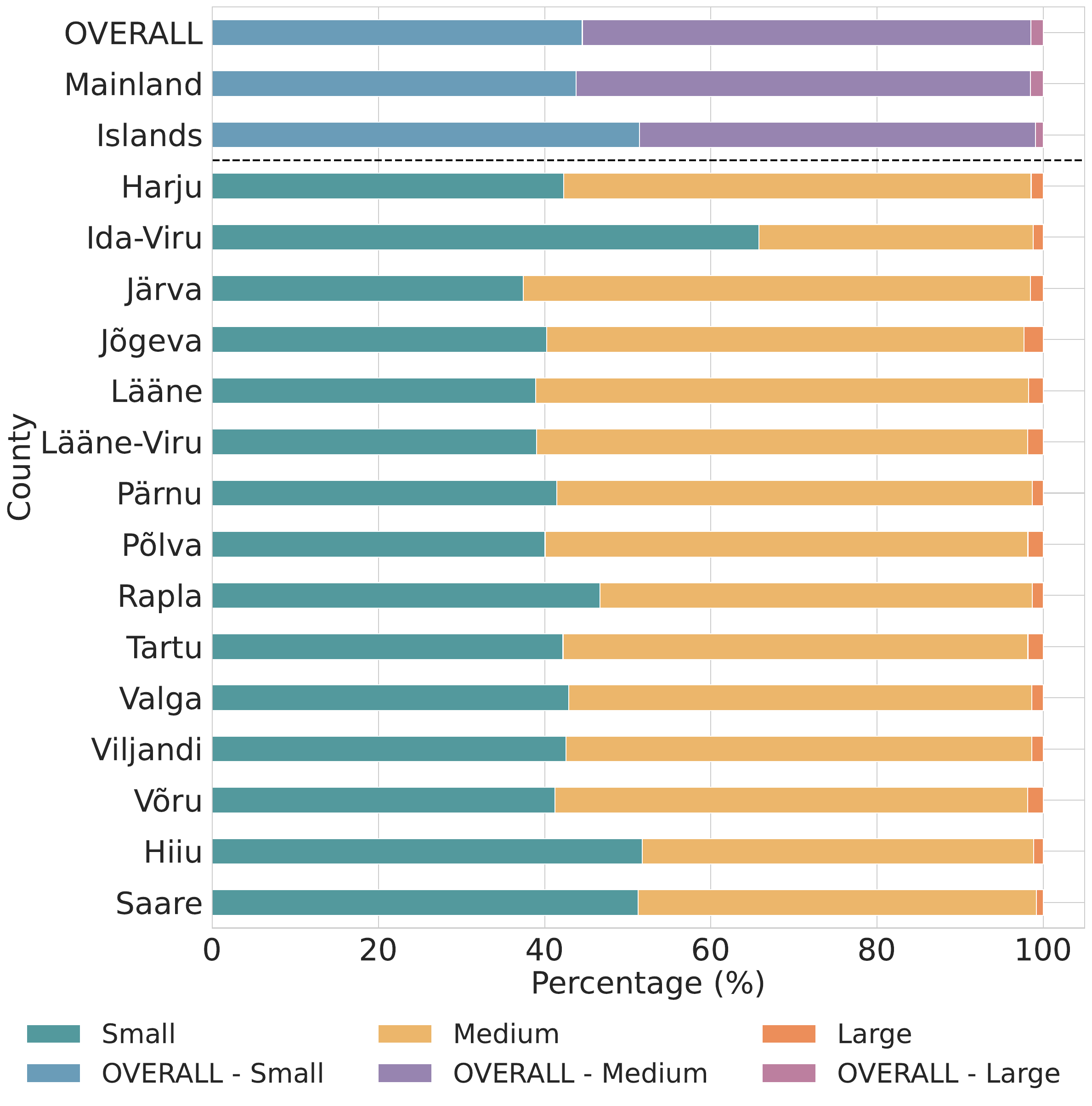}
    \caption{Regional distribution of building scales across Estonian counties. The horizontal scale is categorized into small, medium, and large. The stacked bars illustrate the percentage composition of these categories for each county, as well as aggregated statistics for Mainland, Islands, and the entire dataset (OVERALL).}
    \label{fig:building_scale_stackbar}
\end{figure}

Horizontally, building scales are stratified into three categories including small, medium, and large, following the criteria established by the MS-COCO object detection benchmark~\cite{lin2014microsoft_coco}. Specifically, we mapped pixel-based area thresholds defined by MS-COCO ($\text{Area} < 32^2$ pixels for small, $32^2 \le \text{Area} < 96^2$ for medium, and $\text{Area} \ge 96^2$ pixels for large)~\cite{lin2014microsoft_coco} back to actual physical ground areas based on a predefined projection resolution. Buildings are subsequently categorized based on their real-world footprint area according to these derived physical benchmarks. The resulting regional distribution, visualized as a percentage stacked bar chart in \cref{fig:building_scale_stackbar}, reveals that small and medium-scale structures constitute the vast majority of the dataset, collectively accounting for over \qty{98}{\percent} of the total instances. While most counties exhibit a relatively balanced distribution between small and medium categories, notable regional heterogeneities emerge. For instance, Ida-Viru displays a significantly higher concentration of small buildings (approximately \qty{65}{\percent}), whereas the Islands generally maintain a higher proportion of small-scale units compared to the Mainland. In contrast, large-scale buildings remain a marginal fraction across all counties.

\begin{figure}[t] 
    \centering
    \subfloat[Building Vertices (V)\label{fig:building_vertices}]{
        \includegraphics[width=0.47\columnwidth]{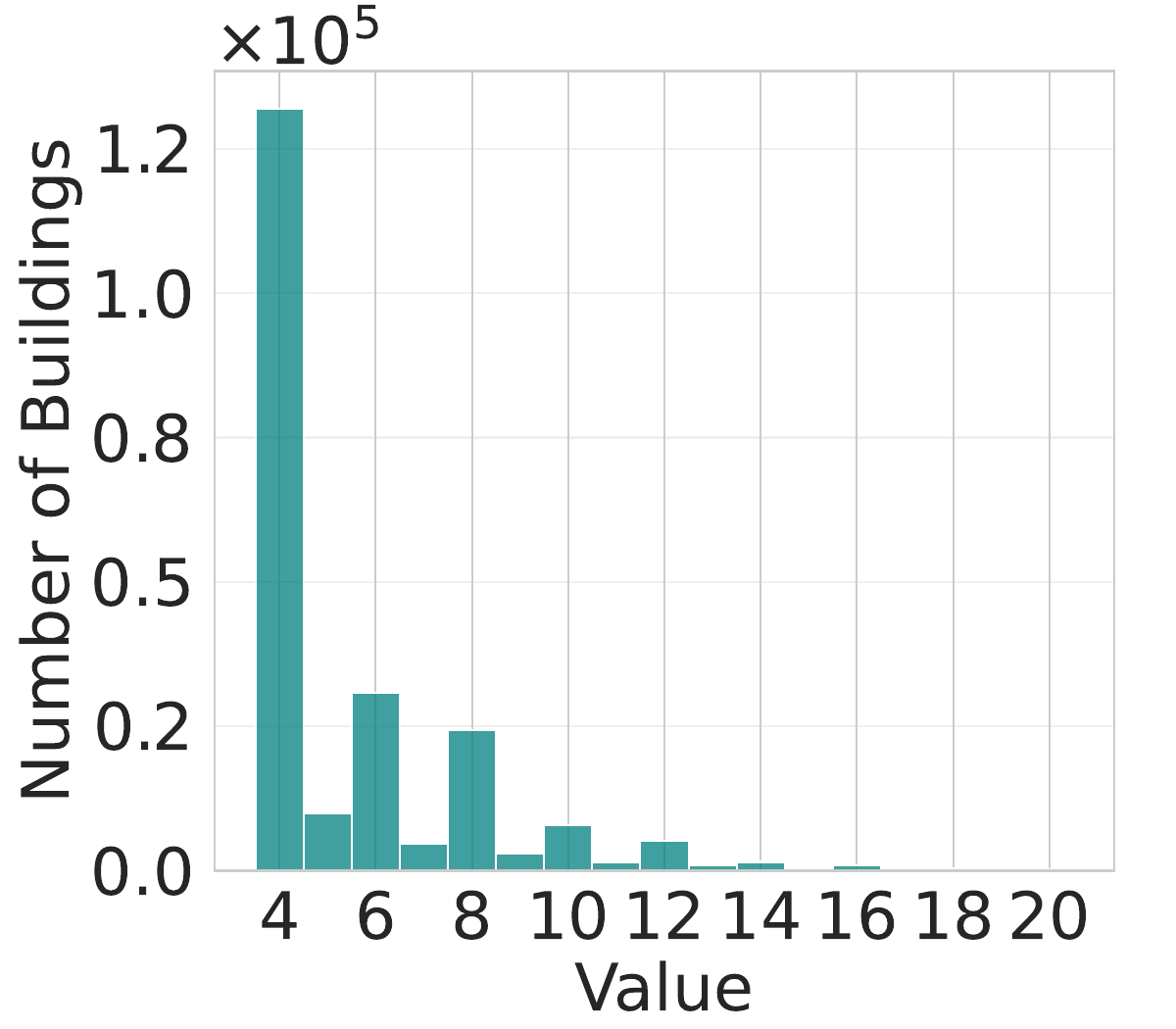}
    }
    \hfill
    \subfloat[Shape Index (SI)\label{fig:shape_index}]{
        \includegraphics[width=0.47\columnwidth]{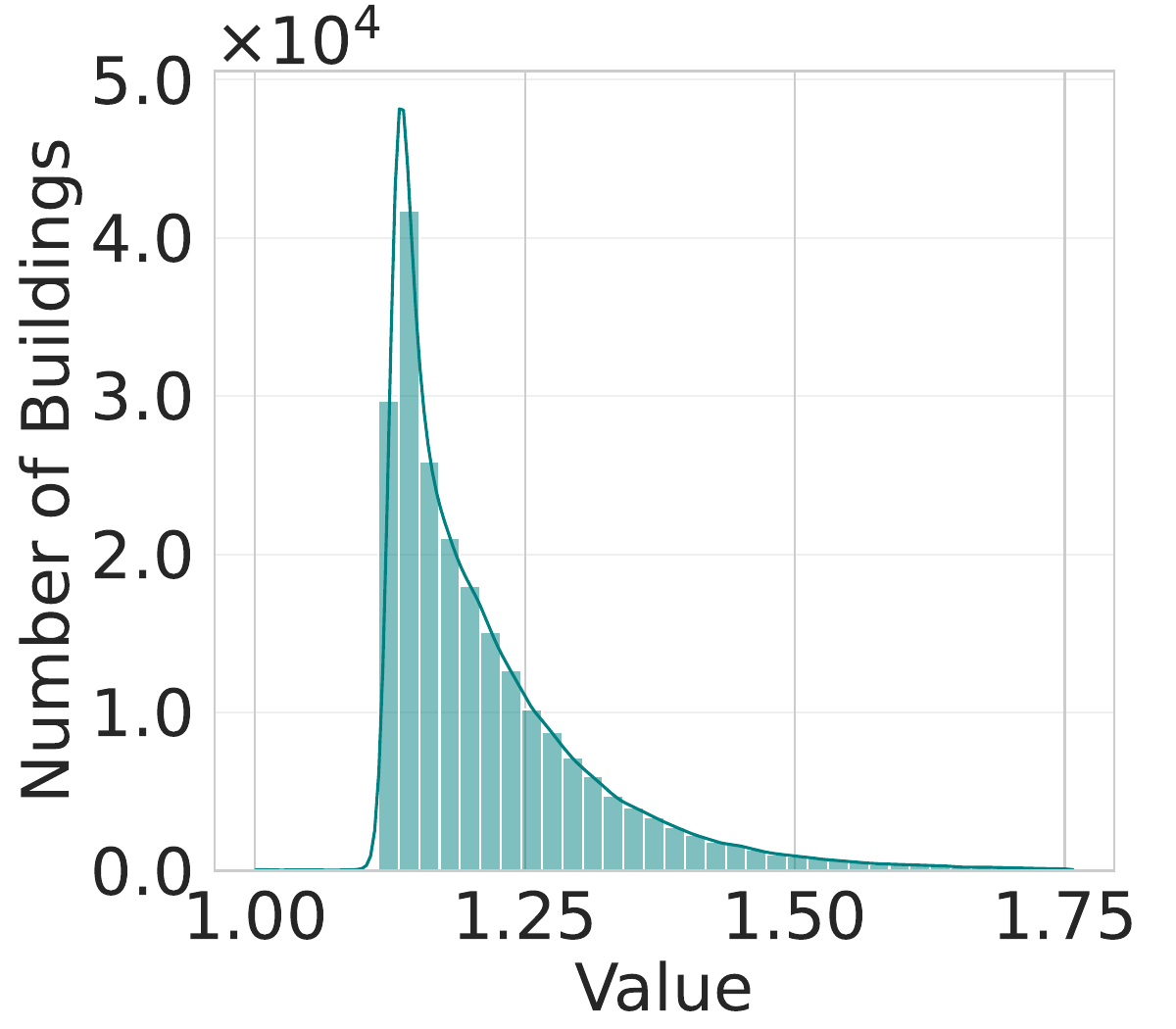}
    }
    \\ 
    \vspace{1mm} 
    \subfloat[Fractal Dimension (FD)\label{fig:fractal_dimension}]{
        \includegraphics[width=0.47\columnwidth]{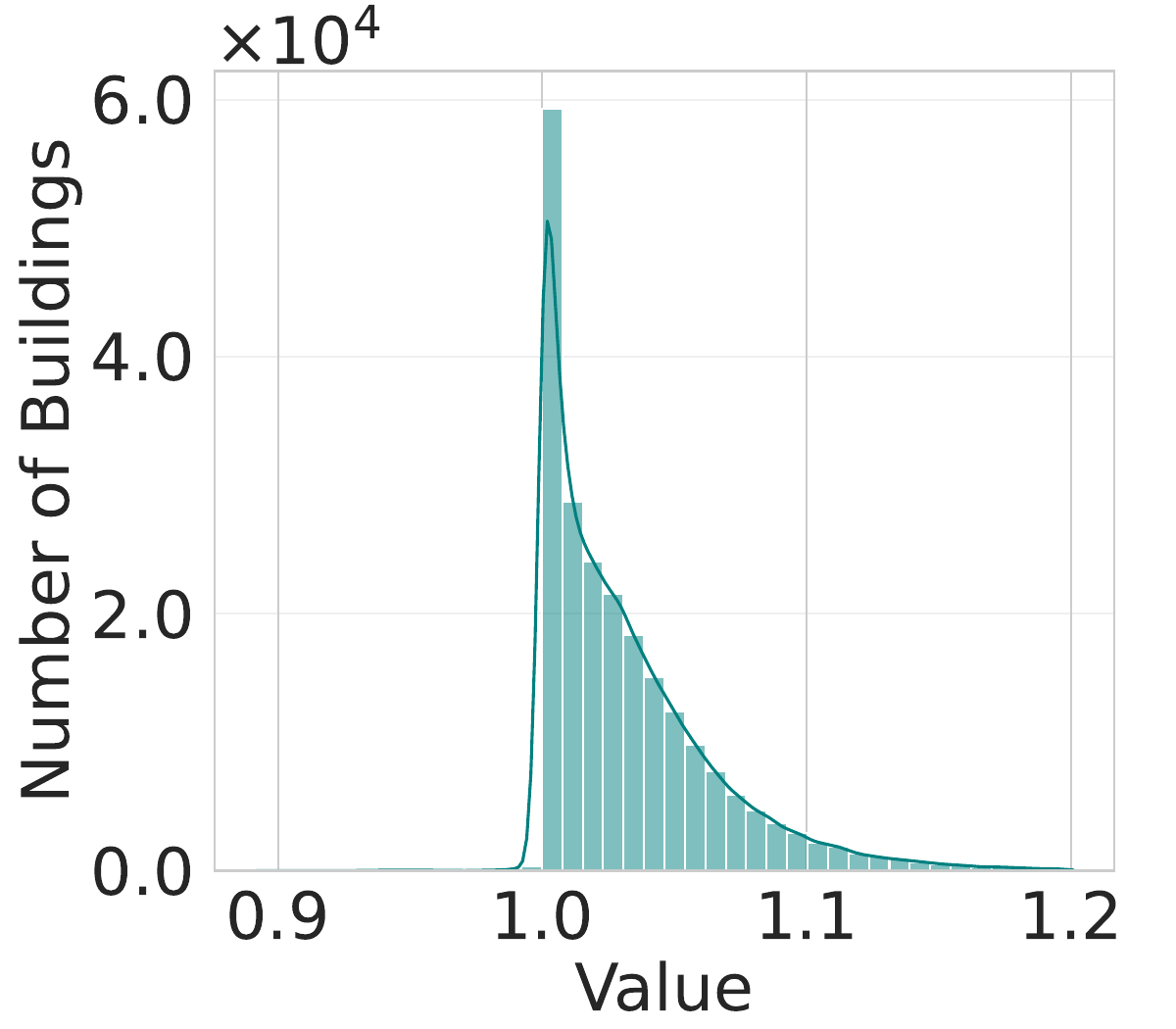}
    }
    \hfill
    \subfloat[Complexity Index\label{fig:complexity_index}]{
        \includegraphics[width=0.47\columnwidth]{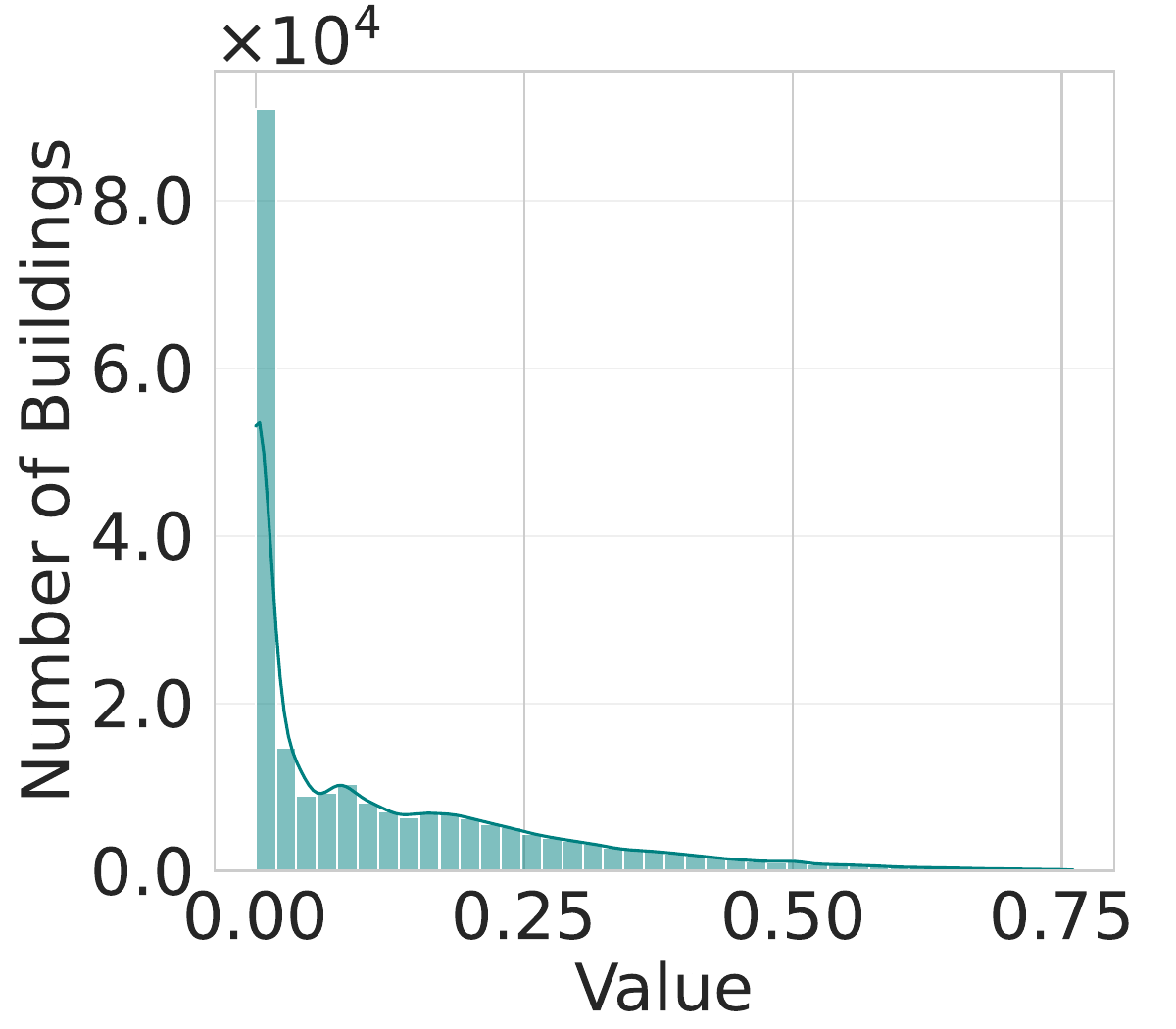}
    }
    \caption{Statistical distribution of building footprint complexity metrics. (a)\mbox{-}(c) Distributions of individual geometric dimensions, including normalized vertex count, shape index, and fractal dimension. (d) Final composite complexity index derived from the weighted combination of the three metrics. The results highlight a right-skewed pattern, indicating that while simple rectangular structures predominate, a significant portion of the dataset consists of irregular and complex geometries.}
    \label{fig:complexity_metrics_distribution}
\end{figure}

\begin{figure}[htbp]
    \centering
    \includegraphics[width=\columnwidth]{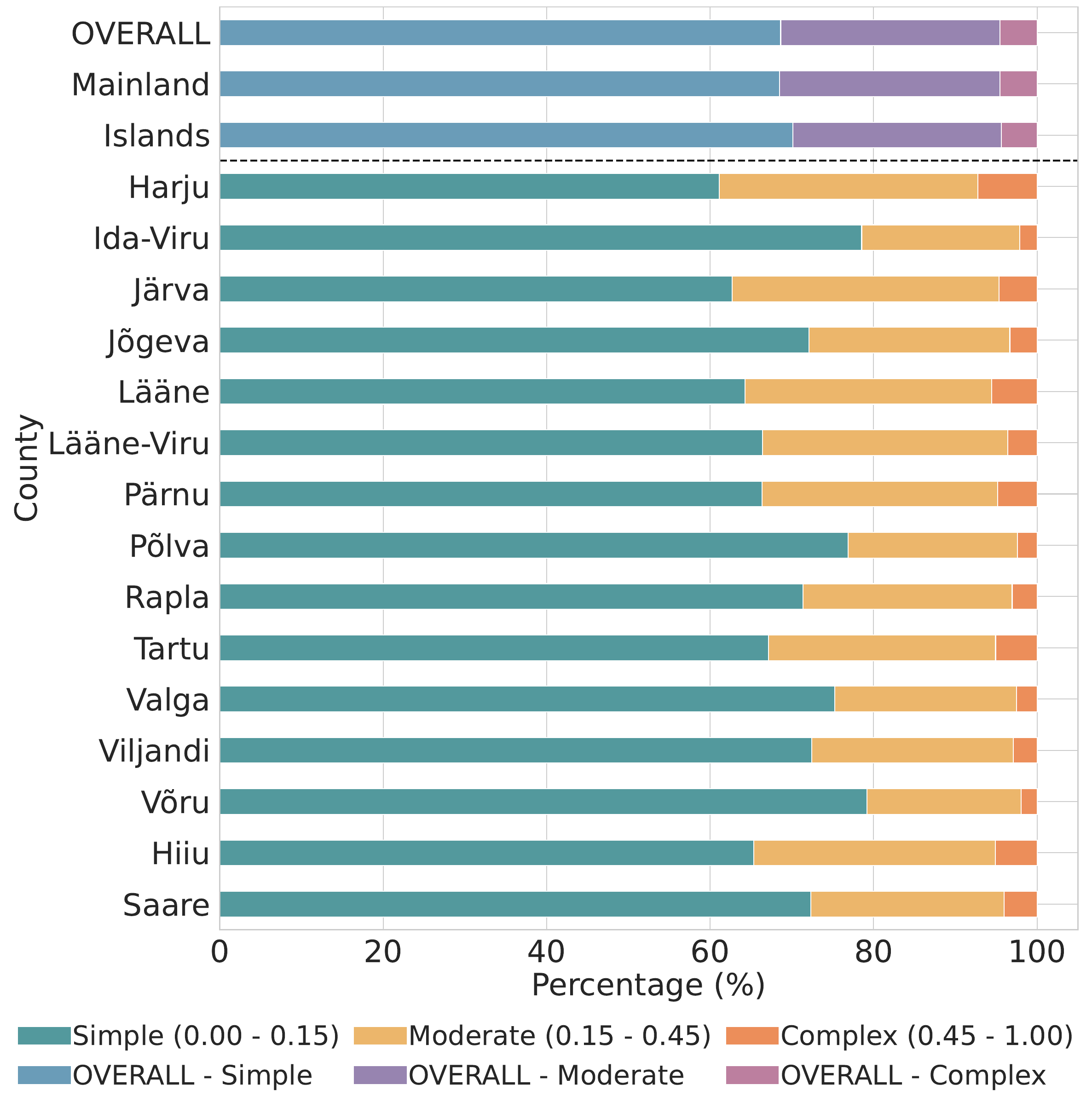}
    \caption{Compositional analysis of building complexity levels across Estonian counties. The horizontal stacked bars illustrate the percentage distribution of simple, moderate, and complex geometries based on the composite complexity index. Aggregated metrics for Mainland, Islands, and the entire dataset (OVERALL) are provided at the top.}
    \label{fig:building_complexity_stackbar}
\end{figure}

\begin{table*}[t]
\centering
\small
\renewcommand{\arraystretch}{1.3} 
\setlength{\tabcolsep}{2pt}
\caption{Comparison of different building footprints extraction datasets and our proposed PCFootprint dataset.}
\label{tab:tab_dataset_comparison}
\begin{tabularx}{\textwidth}{
>{\centering\arraybackslash}m{1.5cm} 
>{\centering\arraybackslash}m{3.2cm} 
>{\centering\arraybackslash}m{2.0cm}
>{\centering\arraybackslash}m{1.6cm} 
>{\centering\arraybackslash}m{1.0cm} 
>{\centering\arraybackslash}m{1.8cm} 
>{\centering\arraybackslash}m{1.9cm} 
>{\raggedleft\arraybackslash}m{0.9cm} 
>{\raggedleft\arraybackslash}m{1.4cm} 
>{\raggedleft\arraybackslash}m{1.3cm}
}
\toprule
\textbf{Data Type} & \textbf{Dataset} & \textbf{Subsets} & \textbf{Vectorized Annotation} & \textbf{Open Source} & \textbf{Elevation Information} & \textbf{Cross-domain Benchmark} & \textbf{Tiles} & \textbf{Building Instances} & \textbf{Coverage [\unit{\km\squared}]} \\ \midrule

\multirow{11}{*}{Imagery} 
& Massachusetts~\cite{MnihThesis} & -- & \colorxm & \colorcm & \colorxm & \colorxm & \num{151} & -- & \num{340} \\
& Inria~\cite{maggiori_inria_2017} & -- & \colorxm & \colorcm & \colorxm & \colorcm & \num{180} & -- & \num{405} \\
& \multirow{2}{*}{ISPRS Dataset~\cite{isprs_semantic_labeling}} & Vaihingen & \colorxm & \colorcm & \colorxm & \colorxm & \num{24} & few & \num{2} \\
& & Potsdam & \colorxm & \colorcm & \colorxm & \colorxm & \num{16} & few & \num{11} \\ \addlinespace[0.5ex]
& WHU-Mix (Raster)~\cite{luo_diverse_2023} & -- & \colorxm & \colorcm & \colorxm & \colorcm & \num{52129} & -- & \num{1213} \\
& \multirow{3}{*}{WHU Building~\cite{ji_WHU_2019}} & Satellite I & \colorxm & \colorcm & \colorxm & \colorxm & \num{204} & -- & -- \\
& & Satellite II & \colorxm & \colorcm & \colorxm & \colorxm & \num{17388} & \num{29085} & \num{550} \\
& & Aerial Imagery & \colorcm & \colorcm & \colorxm & \colorxm & \num{8189} & \num{220000} & \num{450} \\ \addlinespace[0.5ex]
& SpaceNet~\cite{Etten2018SpaceNetAR} & -- & \colorcm & \colorcm & \colorxm & \colorxm & \num{24586} & \num{685235} & \num{5555} \\
& CrowdAI~\cite{mohanty2020crowdai} & -- & \colorcm & \colorcm & \colorxm & \colorxm & \num{341058} & -- & -- \\
& WHU-Mix (Vector)~\cite{wei_buildmapper_2023} & -- & \colorcm & \colorcm & \colorxm & \colorcm & \num{64387} & \num{754126} & \num{1100} \\ \midrule

\multirow{10}{*}{Point Cloud} 
& Fredericton~\cite{vostikolaei_multimodal_2024} & -- & \colorxm & \colorxm & \colorcm & \colorxm & \num{1440} & -- & -- \\ \addlinespace[0.5ex]
& \multirow{5}{*}{Australian~\cite{awrangjeb_automatic_2014}} & Aitkenvale & \colorcm & \colorxm & \colorcm & \colorxm & \num{1} & \num{5} & \multirow{5}{*}{\num{0.236}} \\
& & Hervey Bay & \colorcm & \colorxm & \colorcm & \colorxm & \num{1} & \num{28} & \\
& & Eltham & \colorcm & \colorxm & \colorcm & \colorxm & \num{1} & \num{75} & \\
& & Hobart & \colorcm & \colorxm & \colorcm & \colorxm & \num{1} & \num{69} & \\
& & Knox & \colorcm & \colorxm & \colorcm & \colorxm & \num{1} & \num{52} & \\ \addlinespace[0.5ex]
& Depok~\cite{gamal_semi-automatic_2023} & -- & \colorcm & \colorxm & \colorcm & \colorxm & -- & -- & -- \\
& Hyderabad~\cite{nalini_automatic_2025} & -- & \colorcm & \colorxm & \colorcm & \colorxm & -- & \num{1300} & \num{8} \\ \addlinespace[0.5ex]
\cline{2-10} \addlinespace[0.5ex]
& \multirow{2}{*}{\textbf{PCFootprint (Ours)}} & Mainland & \multirow{2}{*}{\colorcm} & \multirow{2}{*}{\colorcm} & \multirow{2}{*}{\colorcm} & \multirow{2}{*}{\colorcm} & \num{30000} & \num{204608} & \num{491.52} \\
& & Islands & & & & & \num{3000} & \num{22656} & \num{49.15} \\
\bottomrule
\end{tabularx}
\end{table*}

\textbf{Geometric Complexity. } 
To further quantify the geometric morphology of building footprints, we calculate a composite complexity index by integrating three parameters including vertex count ($S_V$), shape index ($S_{SI}$), and fractal dimension ($S_{FD}$). This metric is defined as $Score~=~0.5\cdot S_V + 0.3\cdot S_{SI} + 0.2\cdot S_{FD}$. Specifically, the weighting logic prioritizes vertex count ($S_V$), as it directly determines the presence of a complex primary structural framework. Secondary consideration is given to variability in the aspect ratio via the shape index ($S_{SI}$), as elongated buildings present higher complexity than compact structures with identical vertex count. Finally, the fractal dimension ($S_{FD}$) incorporates boundary roughness to account for minor ancillary structures, though it is assigned the lowest weight to ensure the classification remains focused on the main building geometry. The statistical distributions of these three parameters are visualized in the histograms in \cref{fig:complexity_metrics_distribution}. This reflects the regularity of Estonian building forms, where simple structures predominate while complex, irregular geometries remain a significant minority. Notably, the vertex count distribution reveals a unique parity-based oscillation, where even-numbered vertices (e.g., $4,\ 6,\ 8,\ \dots$) occur with substantially higher frequency than their odd-numbered counterparts. This preference for even vertices indicates a strong structural tendency toward symmetrical or orthogonal designs in the built environment, serving as a critical geometric prior for vectorized reconstruction. Based on the final scores, buildings are stratified into \enquote{Simple} ($[0.00,0.15]$), \enquote{Moderate} ($(0.15,0.45]$), and \enquote{Complex} ($(0.45,1.00]$) levels. Consequently, the regional distribution in \cref{fig:building_complexity_stackbar} reveals a standardized architectural landscape where \enquote{Simple} structures consistently constitute $60\%~\mbox{-}~80\%$ of the building stock. While \enquote{Complex} geometries remain a sparse minority representing less than $5\%$ of the total samples, the high structural consistency between Mainland and Islands aggregates demonstrates a uniform geometric character across Estonia, confirming that while irregular building instances exist, the core urban fabric is dominated by predictable geometric forms. The combination of these structural factors ensures that the PCFootprint dataset provides a rigorous testbed for evaluating the geometric fidelity and spatial robustness of vectorized footprint extraction methods.

\subsection{Dataset Comparison}
To evaluate the positioning of the proposed PCFootprint dataset within the current research landscape, we conducted a comprehensive comparison with existing building footprint extraction benchmarks as summarized in \cref{tab:tab_dataset_comparison}. The comparative analysis focuses on data modality, accessibility, and scale to highlight the unique contributions of this work.

Most established benchmarks for building footprint extraction rely on optical sensors, including high-resolution aerial imagery such as WHU Building~\cite{ji_WHU_2019} and Inria~\cite{maggiori_inria_2017}, or satellite imagery such as SpaceNet~\cite{Etten2018SpaceNetAR} and CrowdAI~\cite{mohanty2020crowdai}. While these datasets~\cite{MnihThesis,maggiori_inria_2017,isprs_semantic_labeling,luo_diverse_2023,ji_WHU_2019,Etten2018SpaceNetAR,mohanty2020crowdai,wei_buildmapper_2023} often feature massive instance counts and are generally open-source, they primarily offer 2D spectral information that lacks the elevation cues essential for precise 3D structural interpretation. Furthermore, as shown in \cref{tab:tab_dataset_comparison}, a majority of existing optical datasets~\cite{MnihThesis,maggiori_inria_2017,isprs_semantic_labeling,luo_diverse_2023} are provided in rasterized annotations, which necessitate additional post-processing to satisfy GIS structural requirements.

Within the point cloud domain, a critical observation from \cref{tab:tab_dataset_comparison} is that most existing ALS datasets specifically annotated for vectorized footprint extraction remain proprietary or restricted, including the Australian Datasets~\cite{awrangjeb_automatic_2014}, the Depok Dataset~\cite{gamal_semi-automatic_2023}, and the Hyderabad Study Area Data~\cite{nalini_automatic_2025}. This lack of accessibility hinders the objective benchmarking of new algorithms and limits the reproducibility of research in footprint extraction from ALS. In contrast, PCFootprint is released as a fully open-source benchmark, aiming to bridge this gap in the community.

Crucially, PCFootprint establishes the first cross-domain benchmark for footprint extraction from ALS point cloud. While certain optical imagery datasets like Inria~\cite{maggiori_inria_2017} and WHU-Mix~\cite{luo_diverse_2023, wei_buildmapper_2023} have successfully incorporated cross-city or cross-regional evaluation, this capability has remained absent in the ALS research domain. Unlike existing point cloud datasets that focus on localized study areas, our benchmark is structured to support both standardized benchmarking workflow and cross-domain evaluation. This is achieved by utilizing tiles from the Estonian mainland alongside a geographically disjoint subset from major islands, enabling the rigorous evaluation of model robustness and zero-shot generalization across varying building morphologies and regional characteristics. 

The scale of PCFootprint significantly exceeds that of existing point cloud benchmarks in spatial coverage and instance count. While the Australian Datasets~\cite{awrangjeb_automatic_2014} and the Hyderabad Data~\cite{nalini_automatic_2025} provide high-fidelity annotations, their coverage is restricted to \qty{0.236}{\km\squared} and \qty{8}{\km\squared}, respectively. PCFootprint offers a total coverage of \qty{540.67}{\km\squared}, which represents an area over \num{60} times larger than the Hyderabad collection~\cite{nalini_automatic_2025} and an order of  magnitude higher than the individual tiles of the Depok~\cite{gamal_semi-automatic_2023} or Australian~\cite{awrangjeb_automatic_2014} series. This extensive coverage encompasses \num{227264} building instances, presenting the high-volume data required for training robust deep learning architectures. By delivering \num{33000} standardized tiles, PCFootprint offers a unified platform for assessing both intra-domain accuracy and the inherent challenges of geographic domain shift.

\begin{figure}[h] 
    \centering
    \includegraphics[width=\columnwidth]{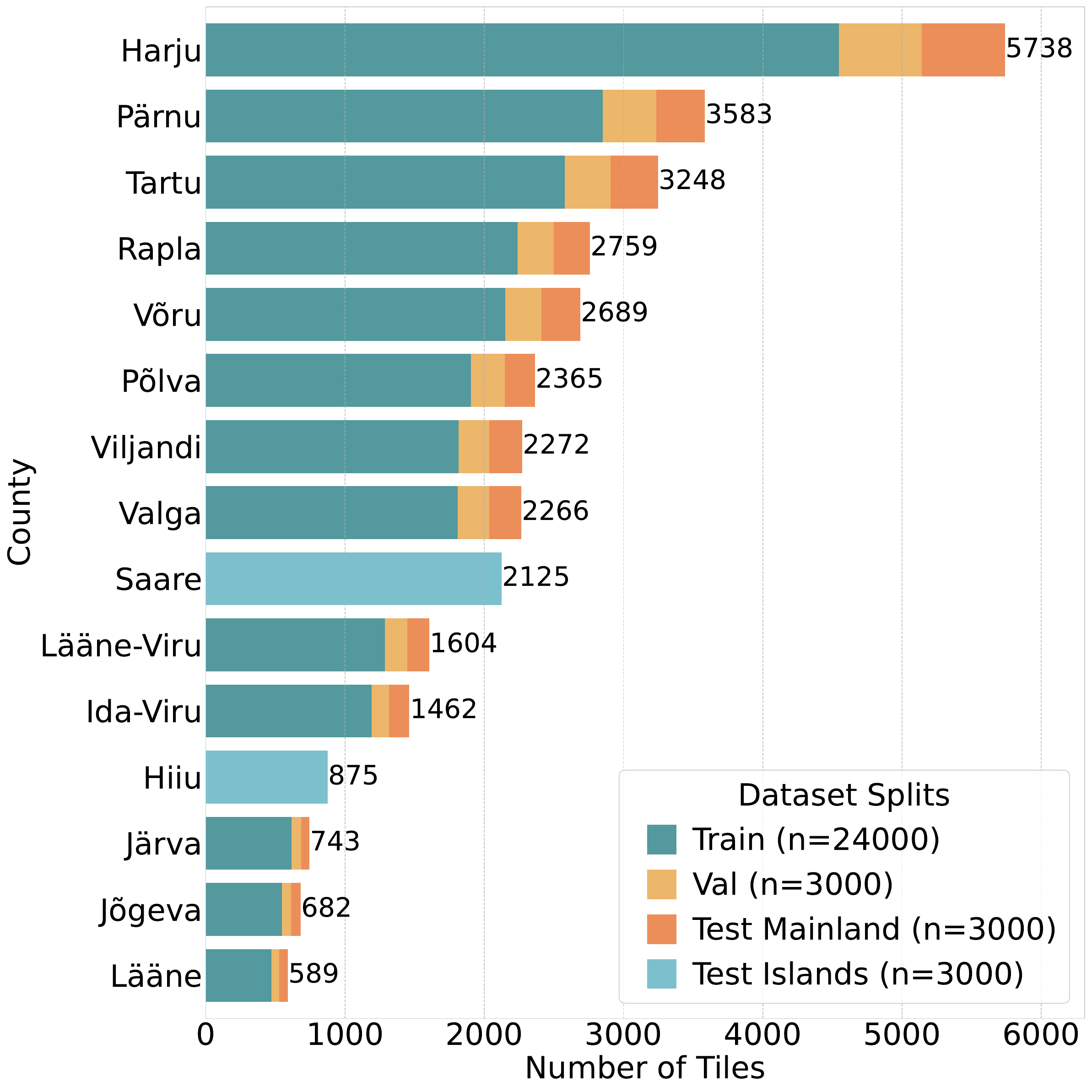} 
    \caption{Proportional distribution and subset splitting of point cloud tiles across Estonian counties. The stacked bars illustrate the allocation for training, validation, and standard benchmark sets within mainland regions, while dedicated Islands subset (Saaremaa and Hiiumaa) constitute the cross-domain generalization benchmark.}
    \label{fig:pointcloud_tiles_split_stackbar}
\end{figure}

\section{Benchmark}
\subsection{Benchmarking Protocols and Data Splitting}
To establish a standardized benchmark for model training and evaluation, the PCFootprint dataset is partitioned into two distinct evaluation protocols: an intra-domain standard benchmark and a cross-domain generalization benchmark. The proportional distribution of tiles across Estonian counties and their allocation into specific subsets are illustrated in \cref{fig:pointcloud_tiles_split_stackbar}.

\begin{enumerate}[label=\arabic*), leftmargin=*]  
\item \textbf{Standard Benchmark.} We designate the Mainland subset as the foundation for our standard benchmark to establish a standardized benchmarking workflow. This protocol utilizes the \num{30000} valid mainland tiles. Extending the density driven sampling strategy used during dataset construction, these tiles were partitioned into training, validation, and testing splits according to a strict $8:1:1$ ratio within each individual county. This stratified approach ensures that the geospatial distribution of all data splits remains synchronized with the actual building density patterns across the Estonian mainland. Specifically, the mainland subset includes \num{24000} samples for training, \num{3000} for validation, and \num{3000} for the standard benchmark set to evaluate in domain performance. While these tiles are strictly non-overlapping to prevent data leakage, they share consistent domain characteristics to facilitate a rigorous assessment of prevailing architectures.
\item \textbf{Generalization Benchmark.} To evaluate the robustness and extrapolation capabilities of algorithms on unseen scenarios, we leverage the Islands subset as the cornerstone of our generalization benchmark. This island-based subset comprises \num{3000} tiles from the major islands of Saaremaa and Hiiumaa, which are geographically isolated from the mainland domain. This cross-domain testbed simulates real-world applications where target inference scenarios often differ significantly from the source training domain in terms of building morphologies and LiDAR sampling patterns. This enables a rigorous assessment of zero-shot generalization performance.
\end{enumerate}
\begin{figure*}[htbp] 
  \centering
  \includegraphics[width=1.0\textwidth]{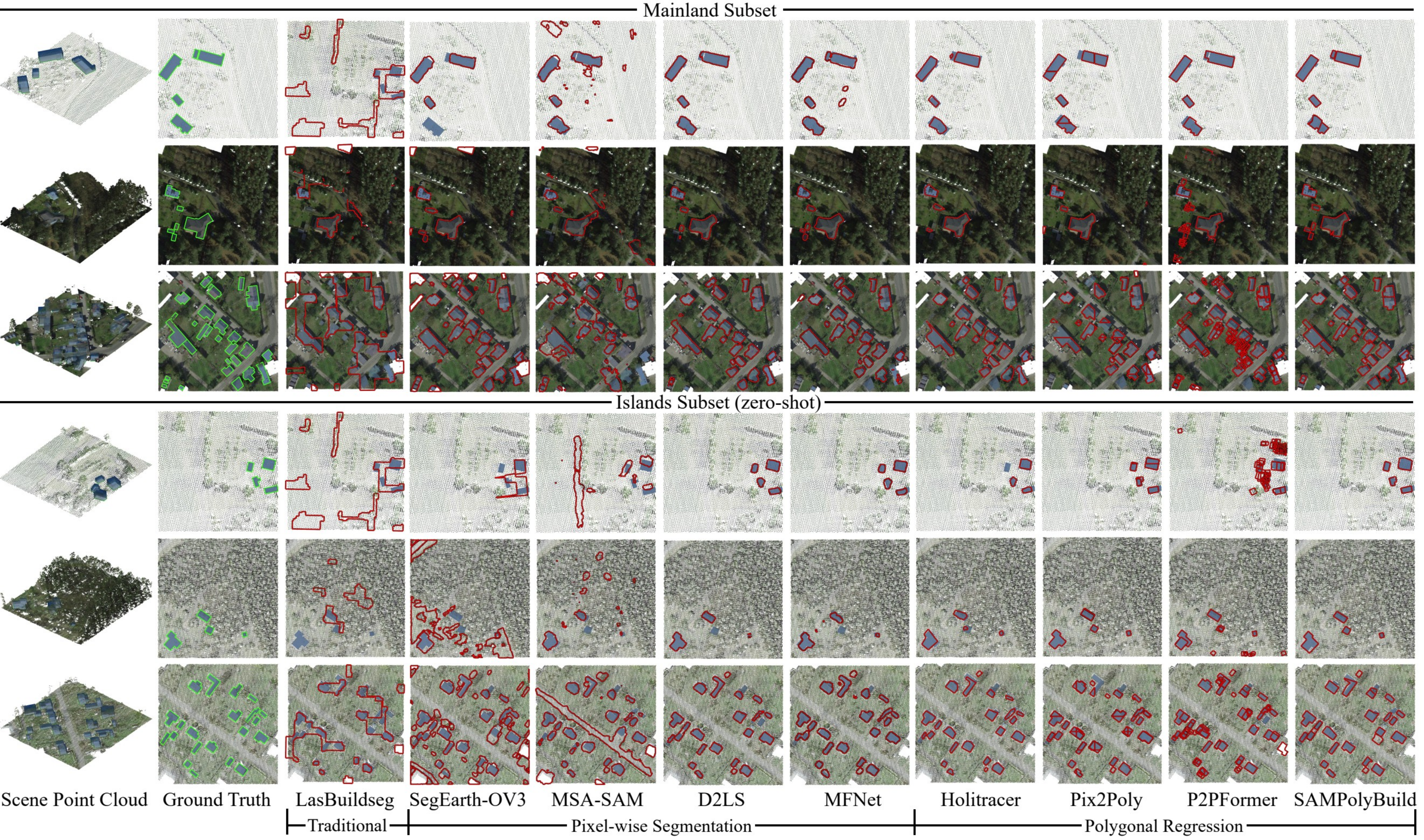} 
  \caption{Qualitative comparison of building footprint extraction results across various methodologies. Representative scenarios characterized by low density sampling, heavy vegetation occlusion, and complex urban morphology were selected to evaluate model robustness. The leftmost column provides 3D visualizations with blue extruded buildings and green ground truth outlines. Subsequent columns present top-view ground truth (second column) and detection results (red outlines). The first three rows represent mainland scenarios, whereas the subsequent three rows display zero-shot generalization results on the islands.}
  \label{fig:results_visualization}
\end{figure*}

\subsection{Evaluation Metrics}

To establish a standardized and objective evaluation of extraction performance, the final predictions from all evaluated models are unified into 2D geometric representations. Regardless of whether the input modality is 3D ALS point clouds or 2D optical imagery, the predicted building footprints are formatted as vectorized polygons or pixel-level masks. This unification allows for the adoption of the MS-COCO evaluation protocol~\cite{lin2014microsoft_coco}, which serves as a rigorous benchmark for instance-level detection and segmentation tasks. By grounding the evaluation in the 2D geometric domain, the framework ensures consistency across distinct model architectures while bypassing the technical complexities of cross-modal alignment during the metric definition stage.

The fundamental criterion for evaluating the spatial correspondence between a predicted footprint and its corresponding ground truth is the Intersection over Union (IoU). 
A prediction is considered a true positive only if its IoU with a ground truth instance exceeds a specified threshold. This mechanism enables the quantification of model performance across varying degrees of localization stringency.

The primary evaluation metrics utilized in this study comprise mean Average Precision ($\text{mAP}$) and Average Recall (\text{AR}). $\text{mAP}$ is calculated by integrating the precision-recall curve across multiple IoU thresholds ranging from \num{0.50} to \num{0.95} with an incremental step of \num{0.05}. This composite metric provides a comprehensive assessment of both detection accuracy and boundary fidelity. We specifically report $\text{AP}_{50}$ and $\text{AP}_{75}$ to represent performance under relaxed and stringent spatial constraints, respectively. Furthermore, $\text{AR}$ is evaluated to measure the model’s ability to successfully recover all building instances within a given scene. Evaluating both $\text{mAP}$ and $\text{AR}$ ensures a balanced analysis of the model’s precision in avoiding false positives and its completeness in mitigating missed detections.

In accordance with the MS-COCO standard~\cite{lin2014microsoft_coco}, the extraction performance is further stratified across different building scales, denoted as $\text{AP}_\text{S}$, $\text{AP}_\text{M}$, and $\text{AP}_\text{L}$. Instances are categorized into Small ($\text{Area} < 32^2$ pixels), Medium ($32^2 \le \text{Area} < 96^2$ pixels), and Large ($\text{Area} \ge 96^2$ pixels) classes based on their pixel area. This multi-scale assessment provides granular insights into the robustness of models, which is particularly critical for the PCFootprint dataset given the significant size variation between small residential structures and massive industrial complexes. By employing this unified and comprehensive evaluation suite, the benchmark facilitates an objective comparison of mainstream algorithms in achieving high-fidelity 2D building footprint reconstruction.

\subsection{Representative Baselines and Results}
\begin{table*}[t]
\centering
\small
\renewcommand{\arraystretch}{1.2}
\caption{Quantitative comparison of building footprint extraction on \textbf{PCFootprint}. Results are grouped by \textbf{Mainland} and \textbf{Islands} subsets and categorized by method {paradigm}. \textbf{Bold} values indicate the best performance for each metric within each subset.}
\label{tab:tab_method_results}
\begin{tabularx}{\textwidth}{>{\raggedright\arraybackslash}m{1.8cm} c @{\extracolsep{\fill}} cccccccc}
\toprule
\textbf{Paradigm} & \textbf{Method} & \textbf{mAP (\%)} & \textbf{AP}$_{\mathbf{50}}$\textbf{ (\%)} & \textbf{AP}$_{\mathbf{75}}$\textbf{ (\%)} & \textbf{AP}$_{\mathbf{S}}$\textbf{ (\%)} & \textbf{AP}$_{\mathbf{M}}$\textbf{ (\%)} & \textbf{AP}$_{\mathbf{L}}$\textbf{ (\%)} & \textbf{AR (\%)} \\
\midrule
\multicolumn{9}{c}{\textit{Mainland Subset}} \\
\midrule
\multirow{4}{=}{Pixel-wise Segmentation} & SegEarth-OV3~\cite{li2025segearthov3} & 8.2 & 21.2 & 4.5 & 1.3 & 14.4 & 15.3 & 28.2 \\
& MSA-SAM~\cite{chen2025msa-sam}     & 15.7 & 40.4 & 7.3 & 1.7 & 20.5 & 17.5 & 21.2 \\
& D2LS~\cite{zou2025dynamic}        & 22.9 & 40.4 & 24.1 & 4.2 & 41.7 & 44.7 & 47.2 \\
& MFNet~\cite{ma2025_mfnet}       & \textbf{49.8} & 76.1 & \textbf{57.8} & \textbf{29.7} & \textbf{62.4} & 56.7 & 54.1 \\
\addlinespace[0.6em] 
\multirow{4}{=}{Polygonal Regression} & HoliTracer~\cite{wang2025holitracer}  & 21.6 & 49.4 & 15.5 & 4.7 & 30.6 & 39.6 & 37.1 \\
& Pix2Poly~\cite{adimoolam2025pix2poly}    & 31.2 & 50.9 & 34.1 & 19.4 & 46.3 & 26.4 & 43.1 \\
& P2PFormer~\cite{zhang2024p2pformer}   & 46.3 & \textbf{77.0} & 50.1 & 17.3 & 52.8 & 60.5 & \textbf{56.7} \\
& SAMPolyBuild~\cite{wang_sampolybuild_2024} & 47.6 & 74.4 & 54.0 & 20.9 & 53.3 & \textbf{66.1} & 54.2 \\
\midrule
\multicolumn{9}{c}{\textit{Islands Subset (zero-shot)}} \\
\midrule
\multirow{4}{=}{Pixel-wise Segmentation} & SegEarth-OV3~\cite{li2025segearthov3} & 1.5 & 4.5 & 0.7 & 0.2 & 4.0 & 0.6 & 18.5 \\
& MSA-SAM~\cite{chen2025msa-sam}    & 11.9 & 33.3 & 4.3 & 1.4 & 15.5 & 16.9 & 16.6 \\
& D2LS~\cite{zou2025dynamic}        & 24.3 & 45.6 & 24.6 & 5.2 & 39.1 & 42.0 & 45.6 \\
& MFNet~\cite{ma2025_mfnet}         & 33.2 & 54.8 & 37.4 & 17.1 & 47.2 & 30.3 & 35.7 \\
\addlinespace[0.6em] 
\multirow{4}{=}{Polygonal Regression} & HoliTracer~\cite{wang2025holitracer}  & 19.8 & 47.7 & 13.0 & 4.2 & 28.2 & 39.5 & 34.1 \\
& Pix2Poly~\cite{adimoolam2025pix2poly}    & 29.9 & 52.7 & 31.4 & 19.4 & 45.0 & 22.2 & 48.2 \\
& P2PFormer~\cite{zhang2024p2pformer}   & 40.8 & 71.1 & 43.1 & 15.8 & 47.0 & 60.2 & 51.3 \\
& SAMPolyBuild~\cite{wang_sampolybuild_2024} & \textbf{44.7} & \textbf{72.3} & \textbf{50.1} & \textbf{21.2} & \textbf{49.9} & \textbf{67.7} & \textbf{51.2} \\
\bottomrule
\end{tabularx}
\end{table*}

To adapt 2D image-based methodologies to the 3D ALS point cloud data, we implemented a standardized projection and interpolation pipeline. Each point cloud tile covering a \qtyproduct{128 x 128}{\m} ground area was projected onto a \numproduct{512 x 512} pixels grid with a ground sampling distance (GSD) of \qty{0.25}{\m} per pixel. The RGB spectral information from the original point clouds was preserved during the projection to form three-channel images. Since the average point density of most tiles is significantly lower than \qty{16}{pts\per\square\metre}, which is the theoretical threshold required to occupy every pixel at a \qty{0.25}{\m} resolution, we utilized the Inverse Distance Weighting (IDW) algorithm for spatial interpolation to generate continuous and dense image representations.

For a rigorous evaluation of prevailing building footprint extraction methods on the PCFootprint dataset, we selected representative baselines encompassing two primary paradigms: pixel-wise segmentation and polygonal regression. Since the deterministic nature of the traditional LasBuildSeg~\cite{erdem2023_lasbuildseg} precludes the generation of probabilistic confidence scores required for MS-COCO~\cite{lin2014microsoft_coco} metrics, this approach is restricted to qualitative assessment to facilitate a holistic comparison of structural fidelity. Quantitative performance across the Mainland and Islands subsets is summarized in~\cref{tab:tab_method_results}, while qualitative visualizations contrasting deep learning baselines with traditional geometric algorithms are presented in~\cref{fig:results_visualization}.

As illustrated in \cref{fig:results_visualization} and quantified in \cref{tab:tab_method_results}, the comparative analysis across representative scenarios including low density sampling regions, areas with heavy vegetation occlusion, and cluttered urban areas reveals distinct performance gaps between traditional heuristics, pixel-wise segmentation, and polygonal regression paradigms. The traditional LasBuildSeg~\cite{erdem2023_lasbuildseg} method performs poorly across all scenarios by generating numerous false positive predictions, arising from the inherent sensitivity of heuristic rules to manual parameter tuning. Similarly, SegEarth-OV3~\cite{li2025segearthov3} and MSA-SAM~\cite{chen2025msa-sam} exhibit significant false detections and omissions. SegEarth-OV3~\cite{li2025segearthov3} achieves a $\text{mAP}$ of only \qty{8.2}{\percent} on the Mainland and drops to \qty{1.5}{\percent} on the Islands. This failure is primarily due to its training-free nature, which relies on SAM 3~\cite{carion2025sam} weights that cannot be fine-tuned for the distinct radiometric and textural characteristics of imagery generated from ALS. MSA-SAM~\cite{chen2025msa-sam} also struggles with robustness, yielding an $\text{mAP}$ of \qty{15.7}{\percent} (Mainland) and frequently misidentifies road surfaces or vegetation as buildings, indicating limited robustness to point cloud derivatives. Despite these localized failures, the relative stability of most models in sparse regions confirms the feasibility of ALS as a reliable source for footprint extraction. Specifically, the pixel-wise segmentation leader MFNet~\cite{ma2025_mfnet} achieves the highest Mainland accuracy (\qty{49.8}{\percent} $\text{mAP}$). Yet pixel-wise segmentation methods exhibit pronounced staircase effect and corner degradation. While these artifacts are less prominent in the Mainland subset, they become highly visible in the Islands generalization set. In contrast, polygonal regression models such as SAMPolyBuild~\cite{wang_sampolybuild_2024} and P2PFormer~\cite{zhang2024p2pformer} achieve the top two results on the Islands with \qty{44.7}{\percent} and \qty{40.8}{\percent} $\text{mAP}$, respectively. They consistently maintain sharp and regularized boundaries across both geographic domains.

Evaluating model generalization necessitates an assessment of performance degradation across the geographically disjoint Islands subset. This dual-testing configuration is vital for identifying the limitations of current architectures under domain shift. When confronted with unseen regional layouts and sampling patterns, the pixel-wise segmentation paradigm exhibits a substantial reduction in $\text{mAP}$ ranging from \num{3.8} to \num{16.6} absolute points. Conversely, the polygonal regression paradigm demonstrates superior spatial robustness by maintaining significantly higher relative precision across distinct geographic domains. The absolute $\text{mAP}$ reduction within the polygonal paradigm is notably more contained, with decreases restricted to between \num{1.3} and \num{5.5} points. This evidence suggests that explicitly modeling geometric primitives is more effective for accommodating variations in building morphology and sampling density than predicting independent pixel labels, thereby providing a more stable solution for zero-shot generalization tasks.

Specific failure modes yield critical insights for future algorithm development. P2PFormer~\cite{zhang2024p2pformer} suffers from severe redundant detections, where multiple overlapping polygons are predicted for a single building instance. This redundancy likely stems from insufficient Non-Maximum Suppression (NMS) or failures in the primitive-matching logic when processing imagery generated from ALS. Similarly, Pix2Poly~\cite{adimoolam2025pix2poly} fails to capture complete building structures and exhibits mild redundancy, suggesting that its post-processing modules are not fully optimized for the non-uniform sampling density of ALS data. Furthermore, the low $\text{AP}_\text{S}$ values across all models indicate a pervasive difficulty in detecting small-scale buildings because sparse point clouds provide insufficient geometric evidence to resolve these structures.

\subsection{Challenges}
The quantitative results and qualitative assessments derived from the PCFootprint benchmark highlight several persistent challenges in vectorized building footprint extraction from ALS point clouds. A primary obstacle involves geographic domain shift, as evidenced by the significant performance discrepancy between the Mainland and Islands evaluation protocols. For instance, prevailing models exhibit a substantial performance degradation, with absolute $\text{mAP}$ reductions ranging from \num{1.3} to \num{16.6} percentage points when transitioning from the Mainland to the Islands subset. This degradation is primarily attributed to variations in architectural styles and land use patterns between the Estonian mainland and the disjoint regions of Saaremaa and Hiiumaa. Such regional heterogeneity constrains the zero-shot generalization capacity of models trained on geographically concentrated data, as they struggle to achieve spatial robustness in unseen urban and rural landscapes.

These regional differences are further compounded by substantial fluctuations in point density, which are fundamentally tied to the heterogeneity of acquisition settings across different survey campaigns. Specifically, varying flight configurations and sensor specifications result in a density range from \qty{0.9}{pts\per\square\metre} to \qty{20.9}{pts\per\square\metre} throughout the PCFootprint dataset. The inherent difficulty of extraction is significantly intensified by the transition from 3D ALS point clouds to 2D image-based workflows, which inevitably sacrifices critical spatial geometric attributes. Since most evaluated methods are image-based, they remain highly sensitive to the quality of 2D projections. In low density regions, the sparsity of the ALS point cloud, a direct consequence of acquisition variability, introduces radiometric artifacts and textural blurring during interpolation. This information loss complicates precise boundary localization particularly for small structures, which is reflected in the pervasive scale sensitivity and lower $\text{AP}_\text{S}$ scores observed throughout the benchmark. Furthermore, image-based methods remain vulnerable to spectral interference and the loss of vertical elevation cues caused by shadows or vegetation occlusion. In complex urban scenes, these limitations frequently cause adjacent building instances to merge. This suggests that a paradigm shift toward the direct exploitation of 3D geometry is essential to achieve the structural integrity and precision required for high-level modeling.

\section{Conclusion}
In this paper, we presented PCFootprint, the first extensive public dataset dedicated to the vectorized extraction of building footprints from ALS point clouds. By providing \num{33000} standardized tiles with systematically aligned annotations, this benchmark fills a critical gap in the field of structured building footprint extraction and cartographic production. Our comprehensive evaluation across diverse geographical landscapes reveals that while pixel-wise segmentation and polygonal regression paradigms achieve significant advancements in geometric fidelity, persistent challenges remain regarding cross-domain generalization. Specifically, the performance degradation observed in geographically disjoint regions underscores the technical necessity for learning domain-invariant geometric features. Furthermore, the limitations inherent in image-based workflows highlight the urgent need for a paradigm shift toward the direct exploitation of 3D spatial geometry. By bridging the gap between raw point clouds and GIS compatible vector outputs, the PCFootprint dataset not only serves as a rigorous testbed for current algorithmic constraints but also paves the way for the next generation of high precision building modeling and digital twin applications.

\section*{Acknowledgments}
The authors thank the Estonian Land and Spatial Development Board for providing access to the raw ALS point cloud data and administrative building records utilized in this study. Appreciation is also extended to the research community for sharing the open-source baseline models and code architectures which were instrumental in the benchmarking process. This work is supported by the Intelligent Computing Center of Shenzhen University.

\bibliographystyle{unsrt}
\bibliography{refs}
\end{document}